\documentclass[pdflatex,sn-mathphys-num,iicol]{sn-jnl}
\usepackage{indentfirst}
\usepackage{graphicx}%
\usepackage{amsfonts}
\usepackage{amssymb}
\usepackage{multirow}%
\usepackage{pifont}

\usepackage{amsmath,amssymb,amsfonts}%
\usepackage{amsthm}%
\usepackage{mathrsfs}%
\usepackage[title]{appendix}%
\usepackage{xcolor}%
\usepackage{textcomp}%
\usepackage{manyfoot}%
\usepackage{booktabs}%
\usepackage{bm}
\usepackage{algorithm}%
\usepackage{algorithmicx}%
\usepackage{algpseudocode}%
\usepackage{listings}%
\usepackage{pdfpages} 
\theoremstyle{thmstyleone}%

\theoremstyle{thmstyletwo}%

\theoremstyle{thmstylethree}%

\begin{document}

\title{Lightweight Multi-Scale Feature Extraction with Fully Connected LMF Layer for Salient Object Detection}

\author*[1]{\fnm{Yunpeng} \sur{Shi}}\email{223800@stu.hebut.edu.cn }
\author[2]{\fnm{Lei} \sur{Chen}}

\author[1]{\fnm{Xiaolu} \sur{Shen}}

\author[1]{\fnm{Yanju} \sur{Guo}}

\affil*[1]{\orgdiv{School of Electronic and Information Engineering}, 
	\orgname{Hebei University of Technology}, 
	\orgaddress{\street{No. 5340 Xiping Road}, 
		\city{Tianjin}, 
		\postcode{300401}, 
		\state{Tianjin}, 
		\country{China}}}
\affil[2]{\orgdiv{School of Information Engineering}, 
	\orgname{Tianjin University of Commerce}, 
	\orgaddress{\street{No. 409 Guangrong Road}, 
		\city{Tianjin}, 
		\postcode{300134}, 
		\country{China}}}

\abstract{In the domain of computer vision, multi-scale feature extraction is vital for tasks such as salient object detection.  However, achieving this capability in lightweight networks remains challenging due to the trade-off between efficiency and performance.  This paper proposes a novel lightweight multi-scale feature extraction layer, termed the LMF layer, which employs depthwise separable dilated convolutions in a fully connected structure.  By integrating multiple LMF layers, we develop LMFNet, a lightweight network tailored for salient object detection.  Our approach significantly reduces the number of parameters while maintaining competitive performance.  Here, we show that LMFNet achieves state-of-the-art or comparable results on five benchmark datasets with only 0.81M parameters, outperforming several traditional and lightweight models in terms of both efficiency and accuracy.  Our work not only addresses the challenge of multi-scale learning in lightweight networks but also demonstrates the potential for broader applications in image processing tasks.
The related code files are available at \href{https://github.com/Shi-Yun-peng/LMFNet}{\texttt{https://github.com/Shi-Yun-peng/LMFNet}}

}


\maketitle

\section{Introduction}

Since AlexNet \cite{krizhevsky2012imagenet} won the ImageNet Large Scale Visual Recognition Challenge (ILSVRC) in 2012, deep neural networks (DNNs) have rapidly evolved, surpassing traditional machine learning methods in accuracy and becoming the dominant approach in computer vision. By stacking multiple convolutional layers, AlexNet enabled the network to learn increasingly complex image features, profoundly influencing subsequent network architectures, such as VGG \cite{simonyan2014very}. However, despite significant performance improvements, DNNs often suffer from an excessive number of parameters and high computational costs, making them challenging to deploy on resource-constrained devices. Moreover, as network depth and complexity increase, performance gains tend to diminish. Consequently, developing efficient neural networks with fewer parameters and reduced computational complexity has become a crucial research direction, driving the growing interest in lightweight network design.  

Optimization strategies for lightweight networks generally fall into two categories: lightweight model design and model compression. Unlike model compression, which reduces redundancy in pre-trained models, lightweight model design fundamentally lowers computational complexity and parameter count, avoiding potential performance degradation caused by compression techniques. Studies have shown that multi-scale feature learning is essential for enhancing model representation capabilities, particularly in dense prediction tasks such as image segmentation and salient object detection (SOD). Traditional convolutional neural networks (CNNs), including VGG and ResNet \cite{he2016deep}, achieve multi-scale feature learning by encoding high-level semantic information in deeper layers while preserving low-level details in shallower ones. However, lightweight networks typically employ fewer layers or reduce the number of channels per layer, limiting the receptive field and impairing their ability to capture multi-scale information in complex scenes. Therefore, a key challenge remains: how to design lightweight networks that effectively learn multi-scale features while maintaining computational efficiency.  

To address this challenge, we introduce a fully connected structure that stacks dilated convolutions with varying dilation rates along different paths, providing diverse receptive fields. Specifically, we propose a simple yet effective multi-scale feature extraction layer, referred to as the Lightweight Multi-scale Feature (LMF) layer. This layer is controlled by a set of hyperparameters that define the number of branches and dilation rates, enabling efficient multi-scale feature learning with minimal computational overhead. Based on the LMF layer, we design a lightweight salient object detection network that achieves competitive performance with only 0.81M parameters. Furthermore, to demonstrate the broader applicability of our approach, we conduct image classification experiments using the same encoder architecture, further validating its effectiveness and generalization ability.
Our main contributions can be summarized as follows:

(1) We propose a lightweight multi-scale feature extraction layer (LMF) based on depthwise separable dilated convolutions, enabling the network to have diverse receptive field variations while maintaining a minimal parameter count.

(2) Using the LMF layer, we design an efficient lightweight network that extracts image features effectively while maintaining a low parameter count.

(3) Extensive experiments on salient object detection and image classification validate the effectiveness and broad applicability of our approach.
\begin{figure*}[ht]
	\vspace{0cm} 
	\centering
	\includegraphics[trim=0.8cm 0.4cm 0.8cm 1.2cm, clip,height=0.32\textheight, width=0.8\textwidth]{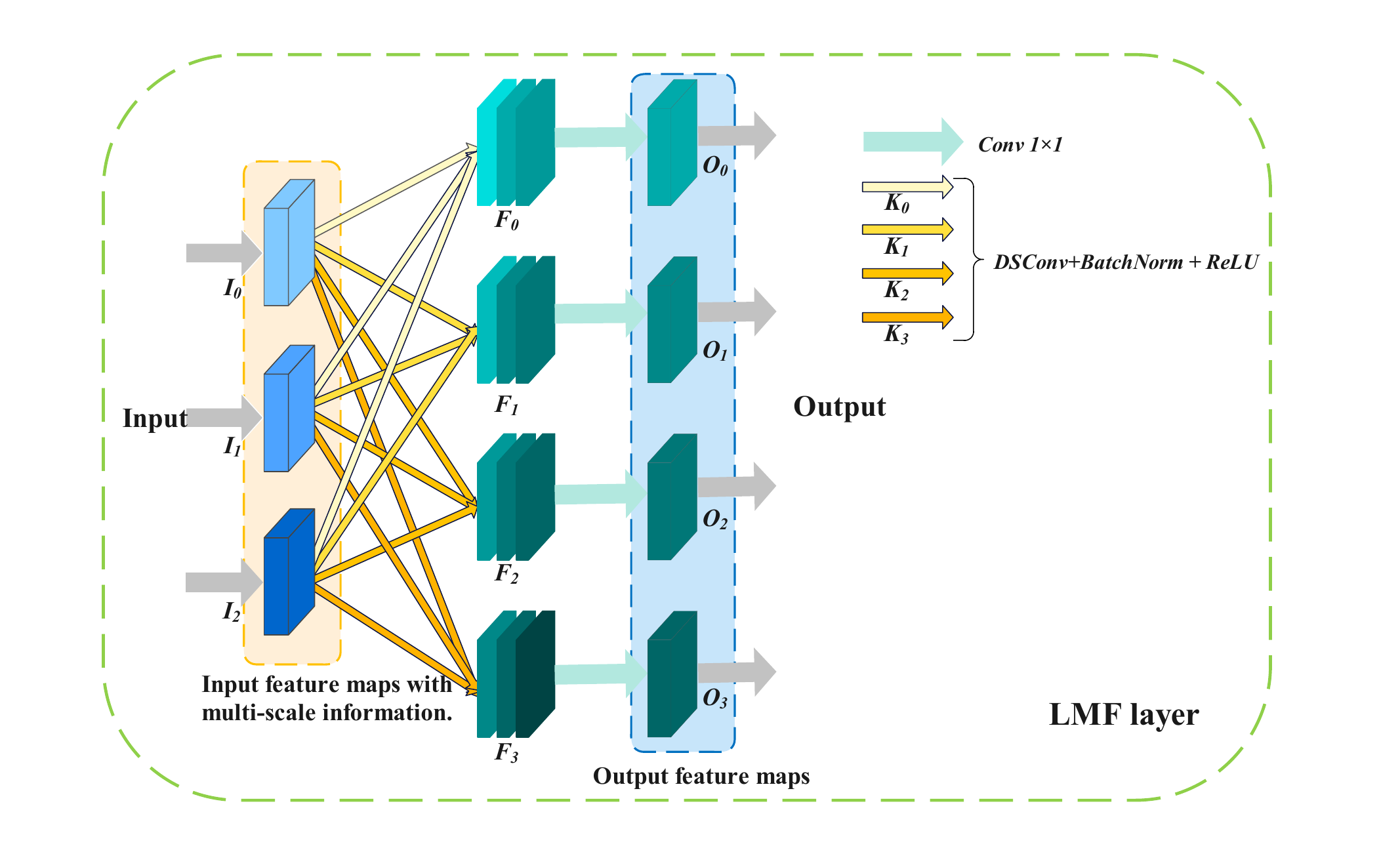}
	\centering
	\caption{Illustration of the LMF layer.  The input feature maps \( \bm{I} = [\bm{I_0}, \bm{I_1}, \bm{I_2}, \dots] \)  contain multi-scale spatial information. Each depthwise separable dilated convolution \( \mathit{K_i} \), with a different dilation rate \( d_i \), processes all inputs to generate intermediate features \( \bm{F_i} \). These are concatenated and fused by a \( 1 \times 1 \) convolution to produce the output feature maps \( \bm{O_i} \). }
	\label{fig:1}
\end{figure*}
\section{Relate Work}

\subsection{Multi-scale Learning}

Multi-scale learning was first introduced by Szegedy et al. in GooLeNet, which employs four parallel branch structures to extract features at different scales \cite{szegedy2015going}. The significant variations in the scale and shape of objects in natural images pose challenges for traditional single-scale methods to effectively model these complexities.  

To address this, researchers have proposed various multi-scale feature learning methods. For instance, Feature Pyramid Networks (FPN) extract and fuse multi-scale features to enhance detection performance, while Atrous Spatial Pyramid Pooling (ASPP) uses atrous convolutions with different dilation rates to expand the receptive field and capture multi-scale information \cite{lin2017feature, chen2017deeplab}.  

The network's ability to learn multi-scale features is closely related to its receptive field. To improve the flexibility of the receptive field, researchers have introduced a series of Deformable Modules. By incorporating learnable offsets into the conventional convolution structure, these modules break the constraints of the fixed sampling grid, allowing the network to better adapt to object deformations and complex backgrounds. However, the high computational cost of Deformable Modules limits their application in lightweight networks.  

Current research on multi-scale learning primarily focuses on efficiently integrating features from different scales \cite{liu2021samnet}, but there is still a lack of exploration into how models can more accurately capture and utilize this information. Therefore, we aim to enhance the diversity of the receptive field in lightweight networks, achieving efficient multi-scale feature extraction while reducing computational overhead.  

To this end, we propose a lightweight multi-scale feature extraction layer based on depthwise separable atrous convolutions. By carefully designing the branch structure, the proposed layer significantly enhances the capability of multi-scale feature modeling while maintaining computational efficiency. Additionally, we incorporate Deformable Modules to further improve the network's adaptability to object deformations. Experimental results demonstrate that our method achieves superior performance in detection and classification tasks, validating its effectiveness and generalization ability.

\subsection{Salient Object Detection}

Salient Object Detection (SOD) is a fundamental task in computer vision aimed at automatically identifying and highlighting the most prominent objects within an image. Over the past few decades, various methods have been proposed, primarily relying on handcrafted features such as contrast and texture\cite{jiang2013salient}\cite{li2013saliency}. With the rapid development of convolutional neural networks (CNNs), deep learning-based SOD methods have become widely adopted for this task\cite{nawaz2020saliency}\cite{liu2018picanet}.

For example, in\cite{qin2020u2}, Qin et al. proposed a nested U-shaped architecture to effectively capture richer contextual information, thereby improving SOD performance. In\cite{ke2022recursive}, Ke et al. introduced a recursive convolutional neural network combined with a contour-saliency blending module, enabling more efficient and accurate salient object detection by exchanging information between contour and saliency features. In\cite{pang2020multi}, Pang et al. utilized an aggregation interaction module to integrate adjacent hierarchical features, achieving promising results.

Despite the strong performance of prior methods, they often require substantial computational resources and storage, making deployment in real-world applications challenging. Existing research suggests that multi-scale and multi-level SOD approaches have shown promising results\cite{liu2021samnet}\cite{pang2020multi}\cite{zhou2024admnet}. Given this, our study focuses on multi-scale feature extraction and proposes a novel lightweight SOD network. Moreover, the output of SOD tasks is typically a binary mask representing pixel-wise classification, which is analogous to image segmentation tasks (i.e., pixel-wise classification)\cite{liu2010learning}\cite{achanta2009frequency}. Therefore, our research may also provide valuable insights and references for other related tasks.

\section{Methodology}

\subsection{Receptive field}
In computer vision, the receptive field refers to the spatial coverage of a filter relative to the source image\cite{kuo2016understanding}. When an image contains multi-scale information, the network typically extracts high-level semantic features in the deeper layers while capturing fine-grained details in the shallower layers\cite{qiu2022a2sppnet}. Therefore, to effectively extract features from larger-scale structures in multi-scale images, the network requires a sufficiently large receptive field. In traditional deep neural network design\cite{he2016deep}, the receptive field is typically expanded by stacking multiple convolutional layers and applying pooling layers.\cite{krizhevsky2012imagenet}\cite{simonyan2014very}
When multiple convolutional layers are stacked, the receptive field for each layer ($RF_i$) can be calculated as follows:
\begin{equation}
RF_i = k\_size_i + (RF_{i-1} - 1) \times d_i
\end{equation}
where $k\_size_i$ is the kernel size, and $d_i$ is the stride.

For example, when $k\_size = 3$, $d = 1$, and pooling layers are not considered, Fig.\ref{fig:2} illustrates how the receptive field changes as three convolutional layers are stacked sequentially. Although stacking convolutional layers increases the receptive field, it also leads to higher computational costs and parameter redundancy, which is inefficient for lightweight neural networks.

\subsection{Depthwise Separable Dilated Convolution}
Depthwise separable dilated convolution combines depthwise separable convolution\cite{howard2017mobilenets} and dilated convolution\cite{chen2017deeplab} to improve computational efficiency and expand the receptive field in CNNs. It first reduces computation and parameters using depthwise separable convolution and then applies dilated convolution to capture broader contextual information without adding extra complexity.

\begin{figure}[htbp] 
	\vspace{0cm} 
	\centering
	\includegraphics[trim=0cm 0cm 0cm 0cm, width=\columnwidth]{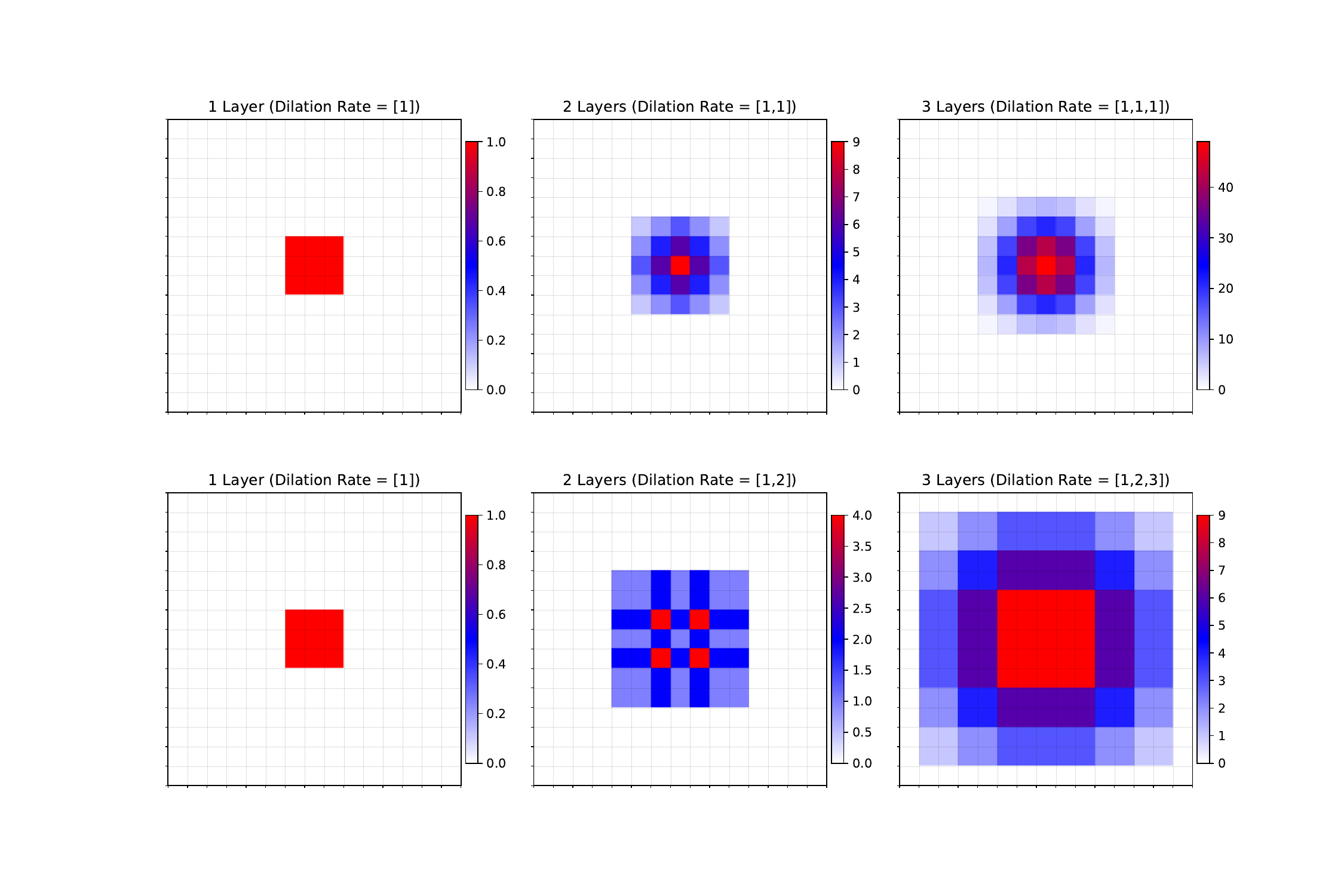} 
	\caption{Illustration of receptive field comparison. The top shows standard convolutions with fixed kernel size, where receptive field growth is gradual. The bottom shows dilated convolutions with increasing dilation rates, significantly enlarging the effective receptive field (in red) by introducing gaps.} 
	\label{fig:2}
\end{figure}

To reduce the number of parameters in convolutional kernels, Andrew G. Howard et al.\cite{howard2017mobilenets} decomposed the traditional convolution operation into depthwise convolution, which processes each input channel independently, and pointwise convolution, which fuses channel information using 1×1 kernels. The number of parameters required for a standard convolution is:
\begin{equation}
\text{params} = k_{\text{size}} \times k_{\text{size}} \times M \times N
\end{equation}
Where \(M\) and \(N\) represent the number of input and output channels of the feature map, respectively.

After decomposing the standard convolution, the number of parameters required is given by:
\begin{equation}
\text{params} = k_{\text{size}} \times k_{\text{size}} \times M + M \times N
\end{equation}
For example, when using a \(3 \times 3\) convolutional kernel with 256 input and output channels, the number of parameters for the standard convolution is 589,824, whereas for the decomposed version, it is 67,840. This decomposition significantly reduces both computational cost and parameter count, with the effect becoming even more pronounced in deeper networks.

Dilated convolution enlarges the receptive field by inserting gaps between kernel elements, enabling the model to capture broader spatial features without increasing kernel size or parameters. As shown in Fig.~\ref{fig:2}, stacking dilated convolutions with rates of 1, 2, and 3 expands the receptive field efficiently, reducing pixel-level redundancy and improving network efficiency.

\subsection{LMF Layer}

Based on the previous analysis, depthwise separable dilated convolutions can expand the receptive field while maintaining the same number of convolutional layers. Stacking layers with different dilation rates results in varying receptive field sizes. To leverage this and capture features at multiple scales, we propose a lightweight multi-scale network layer with a fully connected structure.
﻿
In each layer, depthwise separable convolutions with different dilation rates process the input feature maps. The number and dilation rates are controlled by the dilation factor vector \( \bm{d} = [d_0, d_1, \dots, d_n] \).

Formally, for the LMF layer, the input feature map vector is:
\begin{equation}
\bm{I} = [\bm{I_0}, \bm{I_1}, \bm{I_2}, \dots, \bm{I_m}]
\end{equation}

where \( \bm{I_j} \in \mathbb{R}^{C \times H \times W} \), with \( C \), \( H \), and \( W \) representing the number of channels, height, and width, respectively.Each LMF layer consists of a set of depthwise separable dilated convolutions:
\begin{equation}
\bm{K} = [\mathit{K_0}, \mathit{K_1}, \dots, \mathit{K_n}]
\end{equation}

Each set is controlled by the dilation factor vector \( \bm{d} \).
Formally, for any LMF layer, the input feature map vector is:
\begin{figure*}[h]
	\vspace{0cm} 
	\centering
	\includegraphics[trim=18.5cm 10.5cm 11.5cm 2.5cm, clip,width=1\textwidth]{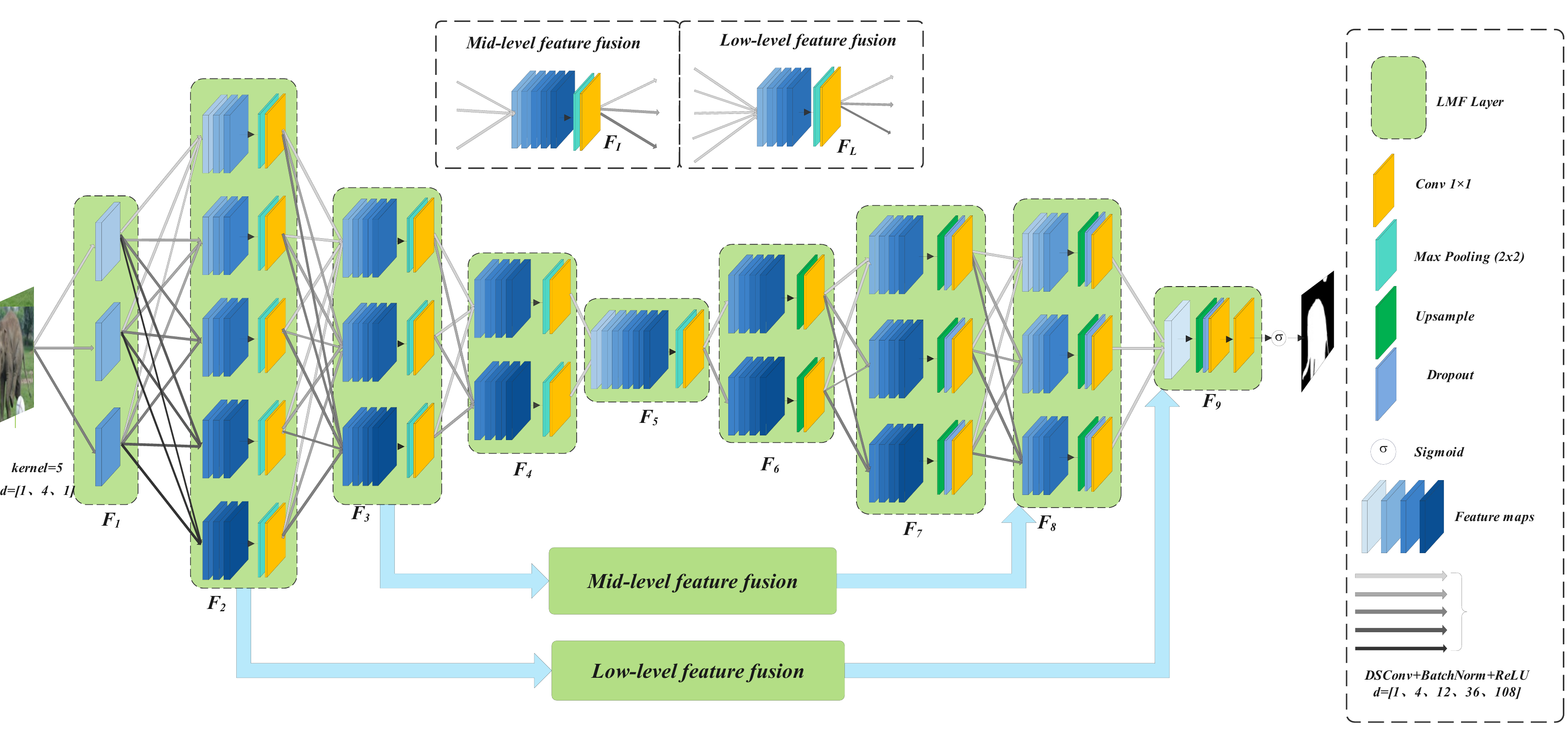}
	\caption{Illustration of the LMFNet architecture. Each LMF layer outputs feature maps \( F_i \) via a \( 1 \times 1 \) convolution. The encoder uses dilated convolutions with increasing rates (\( d_1 = [1,4,12,36,108] \)), with a special configuration for the first layer (\( d_1 = [1,4,1] \)) to reduce information loss and enhance multi-scale feature extraction. The decoder fuses middle and low-level features (\( \mathbf{F}_I, \mathbf{F}_L \)) via bilinear interpolation, followed by Sigmoid activation for the output. }
	\centering
	\label{fig:3}
\end{figure*}
\begin{equation}
\bm{I} = [\bm{I_0}, \bm{I_1}, \bm{I_2}, \dots, \bm{I_m}]
\end{equation}

where \( \mathbf{\mathit{I_j}} \in \mathbb{R}^{C \times H \times W} \), with \( C \), \( H \), and \( W \) representing the number of channels, height, and width of the feature map, respectively.
We use depthwise separable dilated convolutions with different dilation rates as the basic convolution operators in the LMF layer. Therefore, each LMF layer contains a convolution operator vector:
\begin{equation}
\bm{K} = [\mathit{K_0}, \mathit{K_1}, \dots, \mathit{K_n}]
\end{equation}

Each convolution vector is controlled by the dilation factor vector \( \mathbf{d} \).
Let the intermediate feature map vector in the LMF layer be:
\(\bm{F} = [\bm{F_1}, \bm{F_2}, \dots, \bm{F_n}]\)

The intermediate feature map \( F_i \) is represented as:

\begin{equation}
\bm{F_i}= \text{Concat}(\mathit{K_i}(\bm{I_0}), \mathit{K_i}(\bm{I_1}), \dots, \mathit{K_i}(\bm{I_m}))
\end{equation}

After obtaining the intermediate feature maps, we apply a \( 1 \times 1 \) convolution to fuse multi-scale features, producing the output feature vector of the LMF layer.

\begin{equation}
\bm{O} = [\bm{O_1}, \bm{O_2}, \dots, \bm{O_n}]
\end{equation}

Each output feature vector \( \bm{O_i} \) is calculated by:

\begin{equation}
\bm{O_i} = \mathit{K^c}(\bm{F_i})
\end{equation}

Where, \( K^c \) represents the \( 1 \times 1 \) convolution operation used to fuse features from different scales.

\section{Proposed Model}

\subsection{Network structure}
Fig.\ref{fig:3} illustrates the overall network structure incorporating the LMF layer, where \( F_i \) denotes the output feature maps of each LMF layer after a \( 1 \times 1 \) convolution. To enhance practical applicability, we strategically introduce max pooling layers and employ bilinear interpolation for upsampling. 

\begin{figure}[htbp] 
	\vspace{0cm} 
	\centering
	\includegraphics[trim=2cm 11cm 2cm 2cm, width=\textwidth]{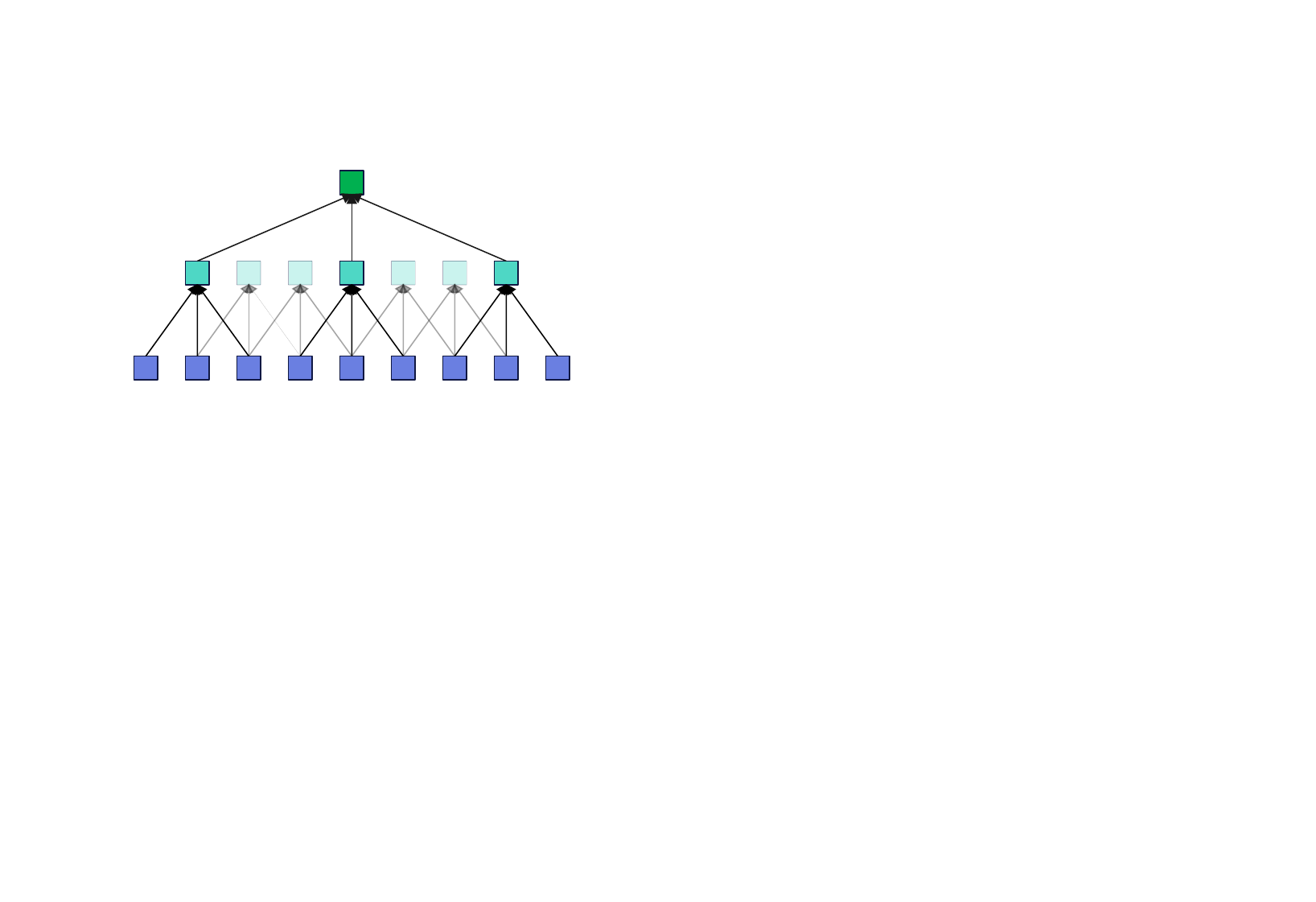} 
	\vspace{0cm} 
	\caption{Illustration of the one-dimensional dilated convolution stacking process. Information loss occurs when the dilation rates of two layers are multiples of each other and exceed the kernel size.} 
	\label{fig:4}
\end{figure}

As shown in Fig.\ref{fig:4}, to mitigate information loss caused by stacking multiple dilated convolutions, we propose that the dilation rate ratio between two adjacent dilated convolution layers should be smaller than the kernel size of the preceding layer. Specifically, for \(3 \times 3\) dilated convolutions, the dilation rate ratio between consecutive layers should not exceed 3. To further prevent information loss due to discontinuous sampling at the lower layers of the network, we set the dilation factors of the first LMF layer to \( d_1 = [1, 4, 1] \). For other LMF layers, the dilation factors are set as \( d_1 = [1, 4, 12, 36, 108] \). When the number of convolution operators is less than 5, the first \( n \) values are selected, where \( n \) denotes the number of convolution operators.

In the encoder, the LMF layer at the bottom contains more dilated convolutions, introducing greater receptive field variations and enabling richer multi-scale feature extraction. At the top, fewer dilated convolutions facilitate multi-scale feature fusion. 
Vertically, the fully connected structure interconnects all dilated convolutions within an LMF layer, forming continuous data pathways. By stacking convolutions with different dilation rates, the network achieves diverse receptive field variations, enhancing multi-scale feature extraction.(Note: Each convolutional layer is followed by batch normalization and an activation function.)

Formally, for the input \(
\mathbf{\mathit{I}} \in \mathbb{R}^{C \times H \times W},\)
in the encoder part, the output of each LMF layer is given by:
\begin{equation}
\left\{
\begin{aligned}
	\mathbf{F}_1 & = f^{LMF_1}(\mathbf{I}), \\
	\mathbf{F}_i & = f^{LMF_P}_i(\mathbf{F}_{i-1}), \quad i = 2, 3, 4, 5,
\end{aligned}
\right.
\end{equation}
Where \( f^{LMF_P}_i \) is the LMF layer with an added pooling layer.

\(\textbf{Mid- and Low-level  Feature Fusion:}\)
Previous studies\cite{liu2021samnet}\cite{zhou2024admnet} have shown that integrating multi-level features can significantly improve network performance. Therefore, following the approach used in the encoder output stage, we employ two LMF layers with max pooling to fuse mid-level and low-level features. In the decoder stage, LMF layers with upsampling adjust the feature maps to the desired dimensions.

Formally, the output feature maps of the feature fusion module are given by:
\begin{equation}
\left\{
\begin{aligned}
	\mathbf{F}_I & = f^{LMF_U}_I(f^{LMF_P}_I(\mathbf{F}_3)), \\
	\mathbf{F}_L & = f^{LMF_U}_L(f^{LMF_P}_L(\mathbf{F}_{2})), 
\end{aligned}
\right.
\end{equation}
Where \(f^{LMF_U}_L\) represents the LMF layer with upsampling.

\(\textbf{Decode:}\) In the decoder, LMF layers with upsampling integrate high-level and low-level semantic information. Finally, a sigmoid layer produces the output. Formally, the feature maps in the decoder are:
\begin{equation}
\left\{
\begin{aligned}
	\mathbf{F}_i & = f^{LMF_U}_i(\mathbf{F}_{i-1}), \quad i = 6, 7, \\
	\mathbf{F}_8 & = f^{LMF_U}_8(Concat(\mathbf{F}_I,\mathbf{F}_7) ), \\
	\mathbf{F}_9 & = f^{LMF_U}_9(Concat(\mathbf{F}_L,\mathbf{F}_8) ), 
\end{aligned}
\right.
\end{equation}
The final output \( S \) is:
\begin{equation}\mathbf{S}=  \mathbf{Sigmod}(f^{Conv}_{1 \times 1}(\mathbf{F}_9 )),\end{equation}

\subsection{Loss Function}
During training, we employ an end-to-end approach, mapping input data directly to the final output without manual design or staged processing. The entire pipeline is optimized holistically, with all parameters updated simultaneously through the objective function. The training process utilizes a hybrid loss function, formulated as:
\begin{equation}
\mathcal{L} = l^{ssim}\left(S, G\right) + l^{bce}\left(S, G\right) + l^{iou}\left(S, G\right)
\end{equation}
where \( G \) represents the ground truth saliency map, and \( S \) denotes the predicted output map of the model. \( \mathcal{L}_{SSIM} \), \( \mathcal{L}_{BCE} \), and \( \mathcal{L}_{IoU} \) correspond to the SSIM loss\cite{wang2003multiscale}, BCE loss\cite{de2005tutorial}, and IoU loss\cite{mattyus2017deeproadmapper}, respectively.

\section{Experiments on SOD}
In this section, we conduct comprehensive experiments on the salient object detection task to validate the effectiveness of the proposed method. Specifically, Section 5.1 provides details on the experimental setup, datasets, and evaluation metrics, while Section 5.2 presents a performance comparison between our model and state-of-the-art approaches.
\subsection{experimental setup}
\subsubsection{Dataset}
We evaluated the proposed method on five widely used public datasets: DUTS\cite{wang2017learning}, ECSSD\cite{yan2013hierarchical}, HKU-IS\cite{li2015visual}, PASCAL-S\cite{li2014secrets}, and DUT-OMRON\cite{yang2013saliency}. These datasets contain 15,572, 1,000, 4,447, 850, and 5,168 natural images, respectively, with corresponding pixel-level annotations. Among them, DUTS is the largest dataset, comprising a training set (DUTS-TR) with 10,553 images and a test set (DUTS-TE) with 5,019 images. Following recent studies\cite{liu2021samnet}\cite{zhou2024admnet}, we trained our model on DUTS-TR and evaluated it on DUTS-TE, ECSSD, HKU-IS, PASCAL-S, and DUT-OMRON.

\subsubsection{Experimental Details}
During data augmentation, we randomly adjust image brightness and contrast, apply random cropping, and perform flipping operations to enhance data diversity and mitigate overfitting. The proposed method is implemented using the PyTorch library. We use the Adam optimizer with \(\beta_1 = 0.9\), \(\beta_2 = 0.999\), and a weight decay of 0.0001. The input resolution is set to \(256 \times 256\), the batch size is 8, and the initial learning rate is 0.001. An exponential learning rate decay strategy with a decay rate of 0.98 is applied. The model is trained for a maximum of 200 epochs, with convergence observed after 146 epochs.

\begin{table*}[ht]
	\centering
	\caption{Quantitative comparison with 10 traditional networks and 5 lightweight networks on the ECSSD, DUTS-TE, HKU-IS, PASCAL-S, and DUT-OMRON datasets. Results inferior to our model are marked in \textcolor{gray}{gray} for traditional networks, and \textcolor{red!90}{red}, \textcolor{green!90}{green}, and \textcolor{blue!90}{blue} are used to mark results for lightweight networks.}
	\resizebox{\textwidth}{!}{%
		\centering
		
		{\fontsize{40}{60}\selectfont
			\begin{tabular}{l|c|c|cccc|cccc|cccc|cccc|cccc}
				\toprule
				\hline
				\multirow{2}{*}{Methods} & \multirow{2}{*}{\/Param (M)} & \multirow{2}{*}{FLOPs(G)} & \multicolumn{4}{c|}{ECSSD} & \multicolumn{4}{c|}{DUTS-TE} & \multicolumn{4}{c|}{HKU-IS} & \multicolumn{4}{c|}{PASCAL-S} & \multicolumn{4}{c}{DUT-OMRON} \\
				\cmidrule(lr){4-23}
				
				& & & MAE $\downarrow$ & {\fontsize{40}{55}\selectfont \textit{F$_\beta$} $\uparrow$} & {\fontsize{40}{55}\selectfont \textit{E$_\beta$} $\uparrow$} & {\fontsize{40}{55}\selectfont \textit{S$_m$}  $\uparrow$} & MAE $\downarrow$ & {\fontsize{40}{55}\selectfont \textit{F$_\beta$} $\uparrow$} & {\fontsize{40}{55}\selectfont \textit{E$_\beta$} $\uparrow$} & {\fontsize{40}{55}\selectfont \textit{S$_m$}  $\uparrow$} & MAE $\downarrow$ & {\fontsize{40}{55}\selectfont \textit{F$_\beta$} $\uparrow$} & {\fontsize{40}{55}\selectfont \textit{E$_\beta$} $\uparrow$} & {\fontsize{40}{55}\selectfont \textit{S$_m$}  $\uparrow$} & MAE $\downarrow$ & {\fontsize{40}{55}\selectfont \textit{F$_\beta$} $\uparrow$} & {\fontsize{40}{55}\selectfont \textit{E$_\beta$} $\uparrow$} & {\fontsize{40}{55}\selectfont \textit{S$_m$}  $\uparrow$} & MAE $\downarrow$ & {\fontsize{40}{55}\selectfont \textit{F$_\beta$} $\uparrow$} & {\fontsize{40}{55}\selectfont \textit{E$_\beta$} $\uparrow$} &
				
				{\fontsize{40}{55}\selectfont \textit{S$_m$} $\uparrow$} \\

				\midrule
				
				AMULet & 33.15 & 45.3 & \textcolor{gray}{0.059} & 0.905 & 0.931 & 0.894 & \textcolor{gray}{0.084} & \textcolor{gray}{0.750} & \textcolor{gray}{0.850} & \textcolor{gray}{0.804} & \textcolor{gray}{0.050} & \textcolor{gray}{0.889} & \textcolor{gray}{0.934} & 0.886 & \textcolor{gray}{0.100} & 0.811 & \textcolor{gray}{0.862} & 0.818 & \textcolor{gray}{0.098} & \textcolor{gray}{0.715} & \textcolor{gray}{0.834} & \textcolor{gray}{0.781}\\

				UCF & 23.98 & 61.4 & \textcolor{gray}{0.069} & \textcolor{gray}{0.890} & \textcolor{gray}{0.922} & \textcolor{gray}{0.883} & \textcolor{gray}{0.111} & \textcolor{gray}{0.741 }& \textcolor{gray}{0.843 }& \textcolor{gray}{0.783} & \textcolor{gray}{0.061} & \textcolor{gray}{0.875} & \textcolor{gray}{0.927} & \textcolor{gray}{0.874} &  \textcolor{gray}{0.116} & 0.792 & \textcolor{gray}{0.847} & 0.805 & \textcolor{gray}{0.120} & \textcolor{gray}{0.698} & \textcolor{gray}{0.821} & \textcolor{gray}{0.760} \\
				RAS & 24.59 & 11.7 & 0.043 & 0.930 & 0.950 & 0.917 & 0.044 & 0.853 & 0.919 & 0.874 & 0.037 & 0.915 & 0.951 & 0.906 & 0.077 & 0.846 & 0.890 & 0.847 & 0.058 & 0.762 & 0.863 & 0.825\\
				
				CPD & 29.23 & 59.5 & 0.040 & 0.923 & 0.944 & 0.910 & 0.043 & 0.839 & 0.911 & 0.867 & 0.033 & 0.911 & 0.950 & 0.905  & 0.072 & 0.842 & 0.888 & 0.845 & 0.057 & 0.747 & 0.856 & 0.818 \\

				BASNet & 87.06 & 127.3 & 0.037 & 0.931 & 0.951 & 0.916 & 0.047 & 0.838 & 0.902 & 0.866 & 0.033 & 0.920 & 0.951 & 0.908 & 0.076 & 0.841 & 0.886 & 0.838 & 0.057 & 0.779 & 0.871 & 0.836 \\

				GateNet & 100.02 & 108.2 & 0.039 & 0.931 & 0.951 & 0.920 & 0.042 & 0.857 & 0.920 & 0.878 & 0.033 & 0.923 & 0.958 & 0.916 &  0.064 & \textcolor{red!90}{0.866} & \textcolor{red!90}{0.910} & \textcolor{red!90}{0.865} & 0.061 & 0.758 & 0.858 & 0.824 \\

				MINet & 47.56 & 146.4 & 0.037 & 0.931 & 0.950 & 0.919 & 0.039 & 0.849 & 0.914 & 0.874 & 0.031 & 0.919 & 0.954 & 0.914 &  0.064 & 0.850 & 0.900 & 0.855 & 0.057 & 0.749 & 0.849 & 0.821 \\

				U2NET & 44.01 & 58.8 & \textcolor{red!90}{0.033} & \textcolor{red!90}{0.941} & \textcolor{red!90}{0.957} & \textcolor{red!90}{0.928} & 0.044 & 0.848 & 0.910 & 0.874 & 0.032 & 0.922 & 0.953 & 0.914  & 0.074 & 0.838 & 0.883 & 0.844 & 0.054 & \textcolor{red!90}{0.793} & \textcolor{red!90}{0.879} & \textcolor{red!90}{0.847} \\

				CDMNET & 26.24 & 18.9 & 0.035 & 0.935 & 0.951 & 0.920 & \textcolor{red!90}{0.033} & \textcolor{red!90}{0.876} & \textcolor{red!90}{0.929} &\textcolor{red!90}{ 0.890} & \textcolor{red!90}{0.028} & \textcolor{red!90}{0.930} & \textcolor{red!90}{0.960} & \textcolor{red!90}{0.919} & \textcolor{red!90}{0.061} & 0.859 & 0.905 & 0.861 & \textcolor{red!90}{0.050} & 0.787 & 0.874 & 0.841 \\
				RCSB & 27.25 & 227.4 & 0.039 & 0.922 & 0.944 & 0.908 & 0.036 & 0.864 & 0.923 & 0.880 & 0.032 & 0.915 & 0.949 & 0.904  & 0.067 & 0.841 & 0.892 & 0.844 & 0.051 & 0.783 & 0.870 & 0.839 \\
				
				LMFNet & \textcolor{red!90}{0.81} & \textcolor{red!90}{3.8} & 0.054 & 0.897 & 0.929 & 0.888 & 0.059 & 0.782 & 0.877 & 0.824 & 0.043 & 0.892 & 0.940& 0.885 &  0.097 & 0.786 & 0.848 & 0.795 & 0.062 & 0.741 & 0.854 & 0.808 \\

				\hline
				& \\[-0.99cm] 
				\hline
				
				CSNet  & \textcolor{red!90}{0.14} & \textcolor{red!90}{0.01} & 0.064 & 0.896 & 0.928 & \textcolor{blue!90}{0.892} & 0.074 & 0.774 & 0.871 & 0.822 & 0.058 & 0.879 & 0.933 & 0.880 & 0.102 & \textcolor{blue!90}{0.797} & \textcolor{blue!90}{0.860} & \textcolor{blue!90}{0.814} & 0.080 & 0.735 & 0.852 & 0.805 \\
				HVPNet & \textcolor{blue!90}{1.23} & \textcolor{blue!90}{1.1}  & \textcolor{green!90}{0.053} & \textcolor{green!90}{0.911} & \textcolor{green!90}{0.939} & \textcolor{green!90}{0.903} & \textcolor{green!90}{0.058} & \textcolor{red!90}{0.814} & \textcolor{green!90}{0.899} & \textcolor{red!90}{0.849} & \textcolor{blue!90}{0.044} & \textcolor{red!90}{0.904} & \textcolor{green!90}{0.947} & \textcolor{red!90}{0.899} & \textcolor{red!90}{0.090} & \textcolor{red!90}{0.816 }& \textcolor{red!90}{0.872} & \textcolor{red!90}{0.829} & \textcolor{green!90}{0.065} &\textcolor{red!90}{ 0.773 }& \textcolor{green!90}{0.875} &\textcolor{red!90}{ 0.831} \\

				SAMNet & 1.33 & \textcolor{green!90}{0.5}  & \textcolor{red!90}{0.051} & \textcolor{red!90}{0.914} & \textcolor{red!90}{0.945} & \textcolor{red!90}{0.907} & \textcolor{red!90}{0.057} & \textcolor{green!90}{0.811} & \textcolor{red!90}{0.901} & \textcolor{green!90}{0.848} & \textcolor{green!90}{0.045} & \textcolor{green!90}{0.903} & \textcolor{red!90}{0.948} & \textcolor{green!90}{0.899} & \textcolor{green!90}{0.092} & \textcolor{green!90}{0.812} & \textcolor{green!90}{0.869} & \textcolor{green!90}{0.825} & \textcolor{blue!90}{0.066} & \textcolor{green!90}{0.772} & \textcolor{red!90}{0.876} & \textcolor{green!90}{0.829} \\
				
				FSMINet & 3.56 & 11.8 & 0.064 & 0.867 & 0.905 & 0.861 & 0.075 & 0.728 & 0.836 & 0.788 & 0.054 & 0.858 & 0.915 & 0.857  & 0.111 & 0.749 & 0.821 & 0.769 & 0.074 & 0.702 & 0.825 & 0.783 \\
				CorrNet & 4.07 & 21.1 & 0.071 & 0.866 & 0.903 & 0.852 & 0.061 & 0.771 & 0.868 & 0.809 & 0.052 & 0.866 & 0.915 & 0.855  & 0.110 & 0.756 & 0.833 & 0.765 & 0.067 & 0.718 & 0.847 & 0.787 \\
				LMFNet  & \textcolor{green!90}{0.81} & 3.8 & \textcolor{blue!90}{0.054} & \textcolor{blue!90}{0.897} &\textcolor{blue!90}{ 0.929} & 0.888 & \textcolor{blue!90}{0.059} &\textcolor{blue!90}{ 0.782} &\textcolor{blue!90}{ 0.877} &\textcolor{blue!90}{ 0.824} &\textcolor{red!90}{ 0.043} &\textcolor{blue!90}{ 0.892} & \textcolor{blue!90}{0.940} & \textcolor{blue!90}{0.885}  & \textcolor{blue!90}{0.097} & 0.786 & 0.848 & 0.795 & \textcolor{red!90}{0.062} & \textcolor{blue!90}{0.741} &\textcolor{blue!90}{ 0.854} & \textcolor{blue!90}{0.808 }\\
				\hline
				\bottomrule
			\end{tabular}%
	}}

\label{tab:1}

\end{table*}

\begin{figure*}[htbp]
	\centering
\begin{minipage}{0.4\textwidth}
	\centering
	\includegraphics[width=\linewidth]{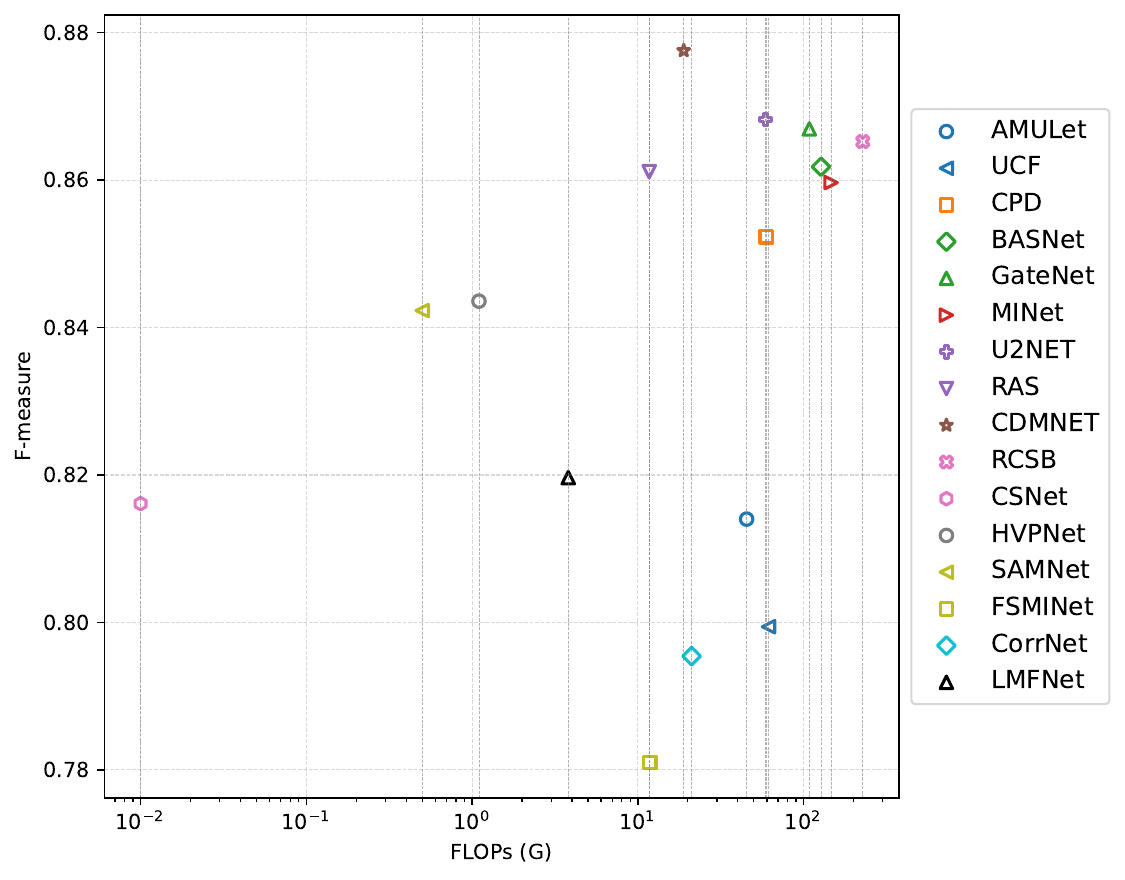}

\end{minipage}
\hspace{0.7cm}
\begin{minipage}{0.4\textwidth}
	\centering
\includegraphics[width=\linewidth]{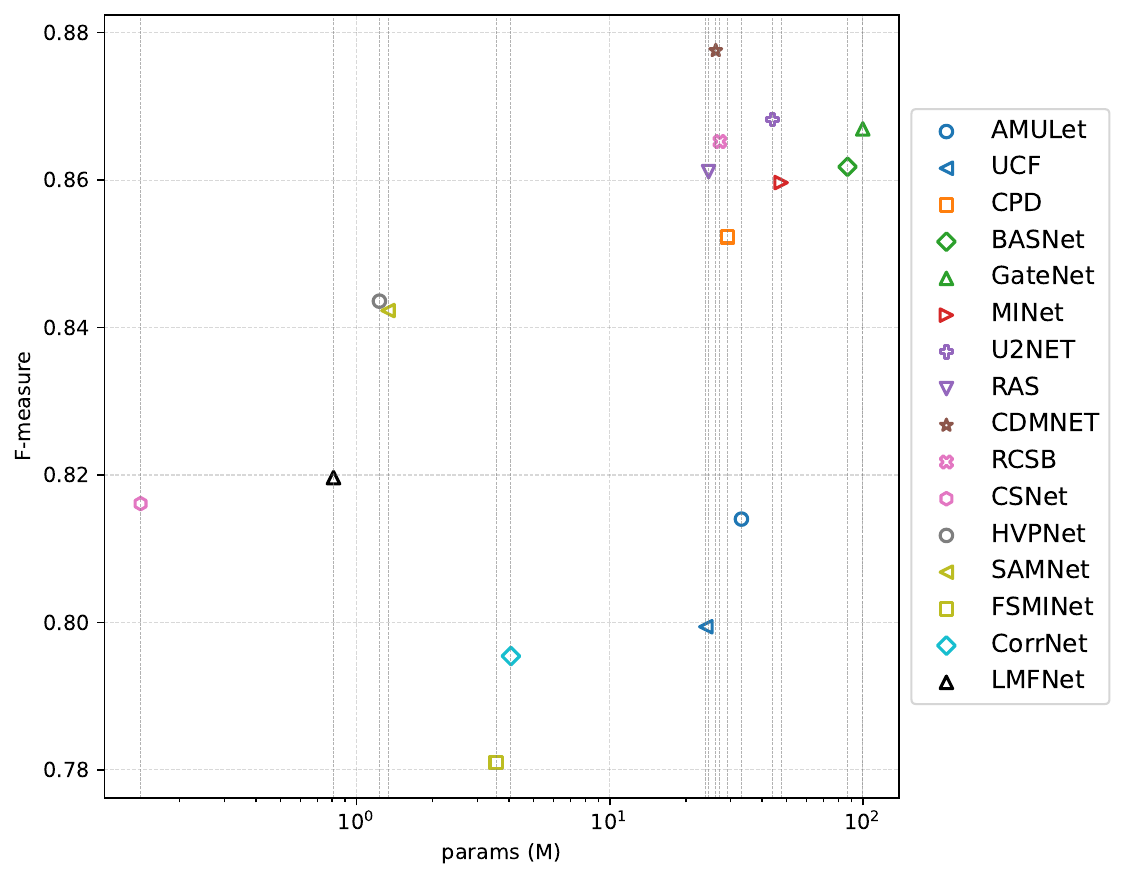} 
\end{minipage}
\caption{Illustration of the tradeoff between performance and computational cost. The F-measure value is averaged over five datasets.Note that the horizon axis is logarithmic.}
\label{fig:5}
\end{figure*}

\subsubsection{Evaluation Metrics}
In this task, we use five widely accepted evaluation metrics: Mean Absolute Error (MAE), Maximum F-measure (F\(\beta\))\cite{achanta2009frequency}, Maximum E-measure (E\(\beta\))\cite{fan2018enhanced}, Adaptive S-measure(Sm)\cite{fan2017structure}, Precision-Recall Curve (PR curve), and F-measure Curve. Among these, MAE measures the average absolute error between the predicted feature map \( S \) and the ground truth \( G \) at each pixel, which is formally expressed as:
\begin{equation}
\text{MAE} = \frac{1}{W\times H} \sum_{i=1}^{W\times H} |S_i - G_i|
\end{equation}
where \( S_i \) and \( G_i \) represent the pixel values of the predicted feature map and ground truth at the \( i \)-th pixel, respectively.
﻿
The F-measure is calculated by computing the weighted harmonic mean of precision and recall, and is given by the following formula:
\begin{equation}
F_{\beta} = \frac{(1 + \beta^2) \cdot \text{Precision} \cdot \text{Recall}}{\beta^2 \cdot \text{Precision} + \text{Recall}}
\end{equation}
According to\cite{liu2021samnet}\cite{zhou2024admnet}, we set \( \beta^2 \) to 0.3. In our evaluation, we report the maximum F-measure value and plot the corresponding F-measure curve. Specifically, the F-measure curve is generated by calculating the F-score for each threshold in the range of [0, 255], as shown in formula (15).

E-measure is used to evaluate the similarity between the saliency map and the ground truth by considering both the local pixel saliency values and the overall image-level average saliency values.

S-measure quantifies the structural similarity of the predicted saliency map, which is defined by combining region similarity (\( S_r \)) and object similarity (\( S_o \)), as shown in the following formula:
\begin{equation}
S_m = \alpha S_r + (1 - \alpha) S_o
\end{equation}
﻿
Where \( S_r \) represents the region similarity, which measures the similarity between the predicted saliency map and the ground truth at the region level, and \( S_o \) represents the object similarity, which evaluates how well the predicted saliency map aligns with the ground truth at the object level. The parameter \( \alpha \) controls the trade-off between these two factors, typically set to 0.5 for balanced consideration of both region and object similarities.

Additionally, similar to\cite{liu2021samnet}and\cite{liu2020lightweight}, we use floating-point operations (FLOPs) and the number of model parameters (Params), measured in millions (M) and gigabytes (G), to evaluate the model's efficiency. Specifically, "Params" refers to the total number of parameters in the model, which serves as an indicator of storage resource consumption. "FLOPs" refers to the floating-point operations count, used to quantify the computational complexity of the network.

\begin{figure*}[h]
	\newgeometry{left=0.8in, right=0.8in} 
	\centering  
	
	\includegraphics[trim=0cm 0cm 0cm 0cm, clip, width=1\textwidth]{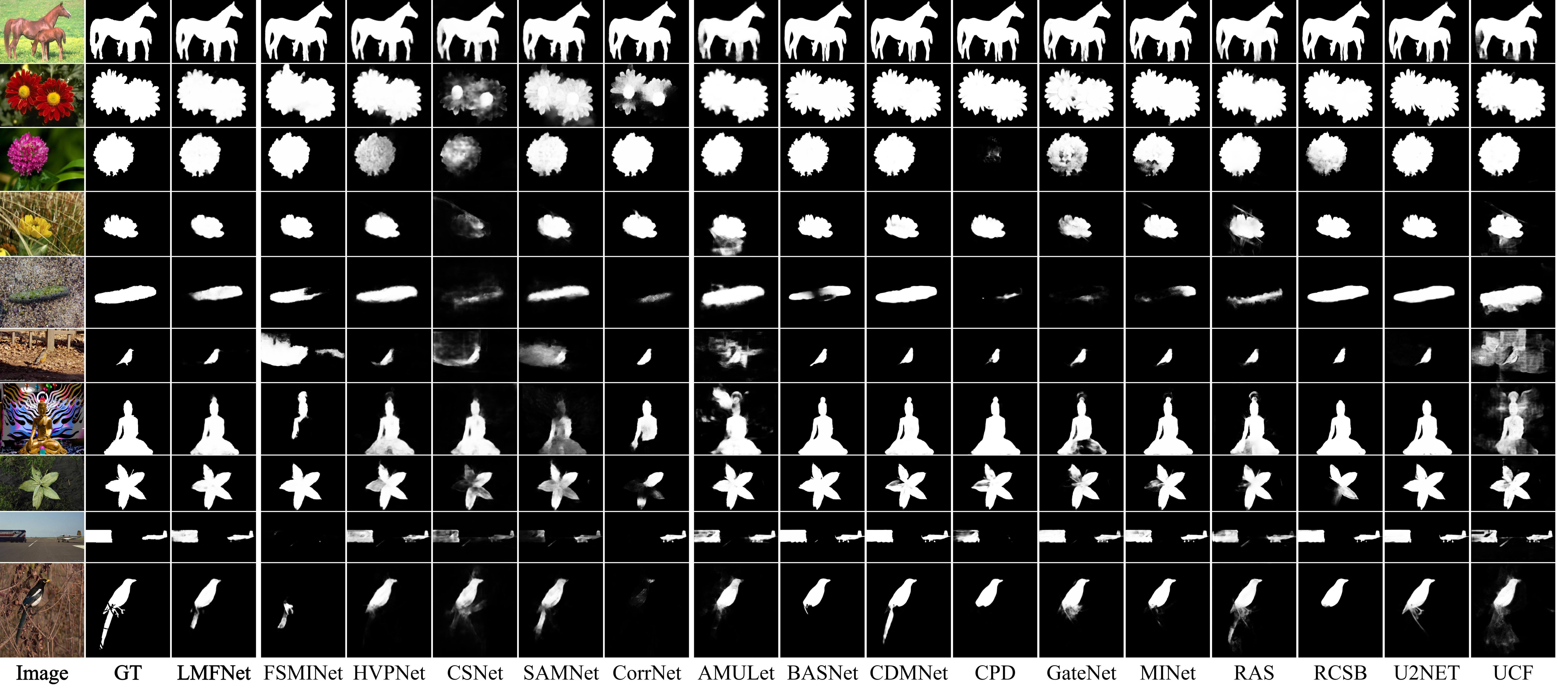}
	\caption{Visual comparison of different saliency object detection methods.}
	\label{fig:6}
	
	\vspace{0.5cm}
	
	\centering
	\begin{minipage}{0.162\textwidth}
		\centering
		\includegraphics[width=\linewidth]{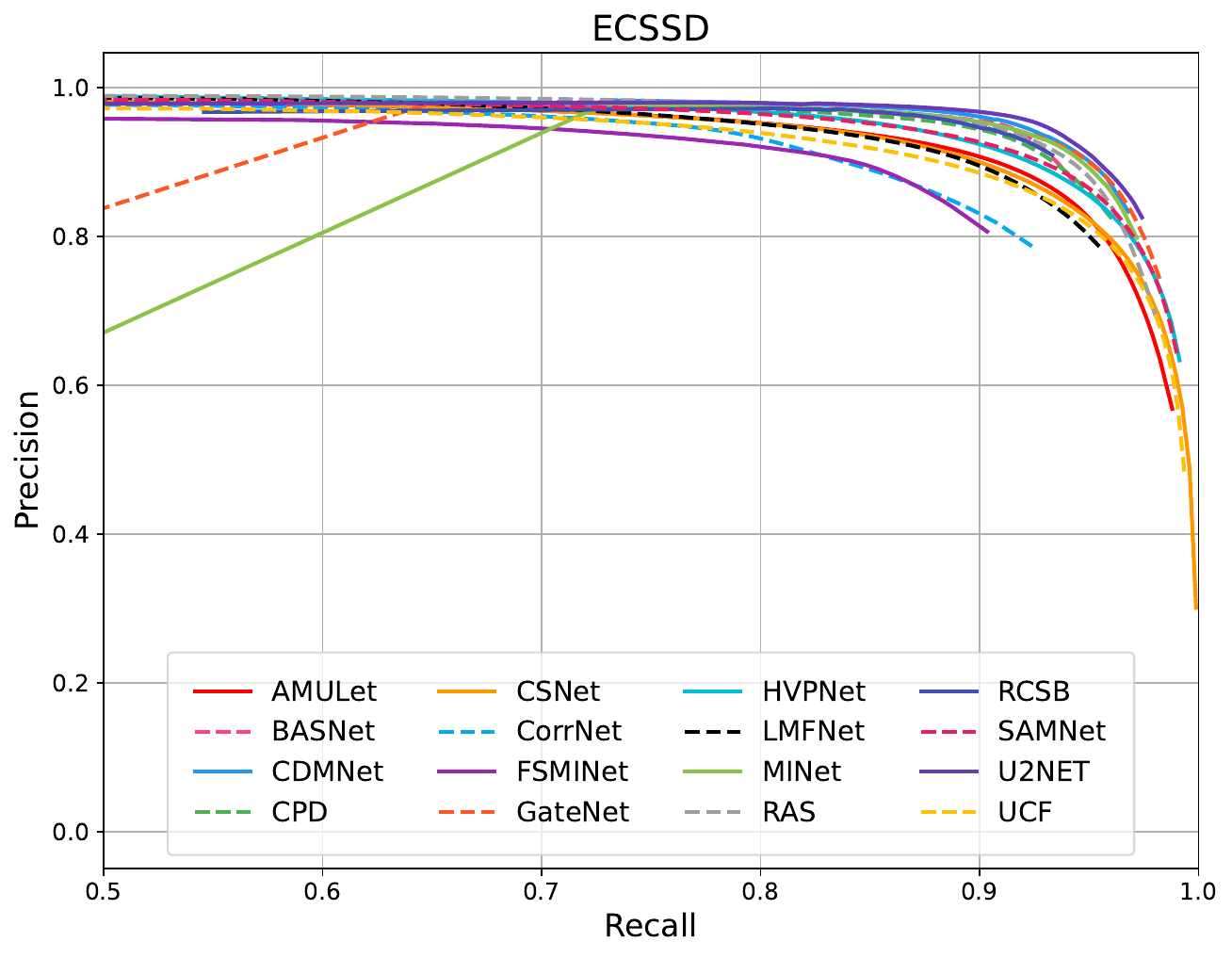} \\[1ex]
		\includegraphics[width=\linewidth]{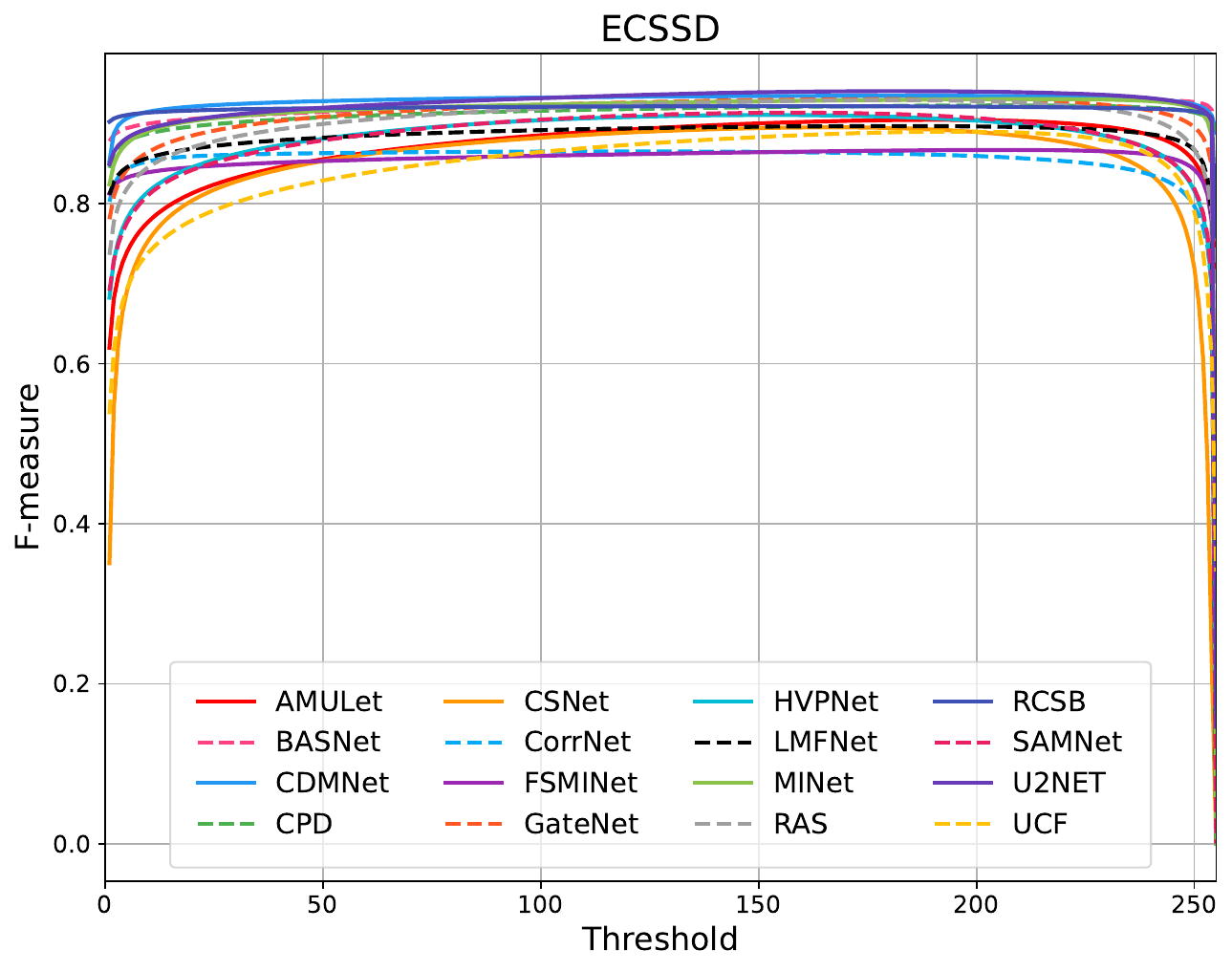} \\[1ex]
		ECSSD
	\end{minipage}
	\hfill
	\begin{minipage}{0.162\textwidth}
		\centering
		\includegraphics[width=\linewidth]{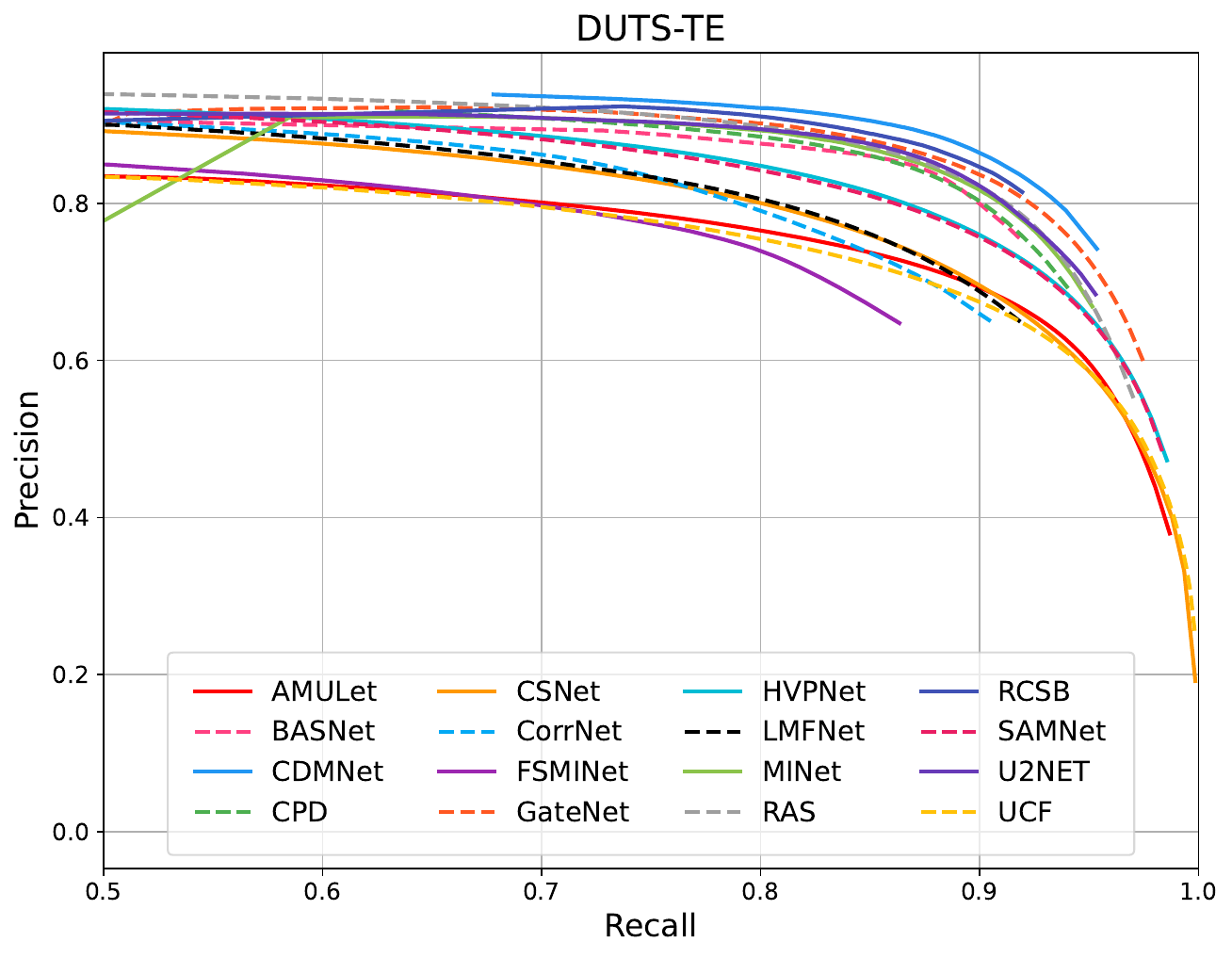} \\[1ex]
		\includegraphics[width=\linewidth]{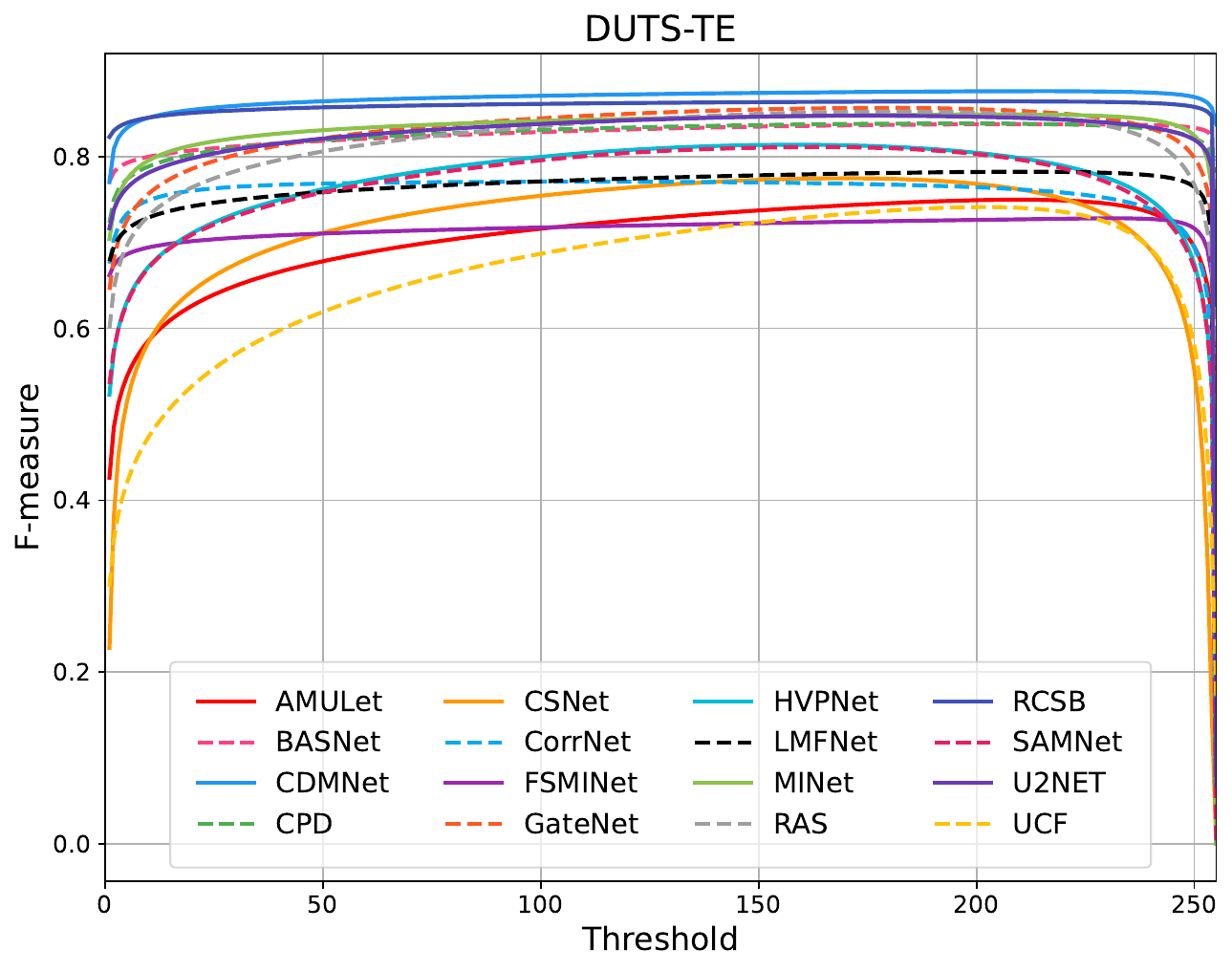} \\[1ex]
		DUTS-TE
	\end{minipage}
	\hfill
	\begin{minipage}{0.162\textwidth}
		\centering
		\includegraphics[width=\linewidth]{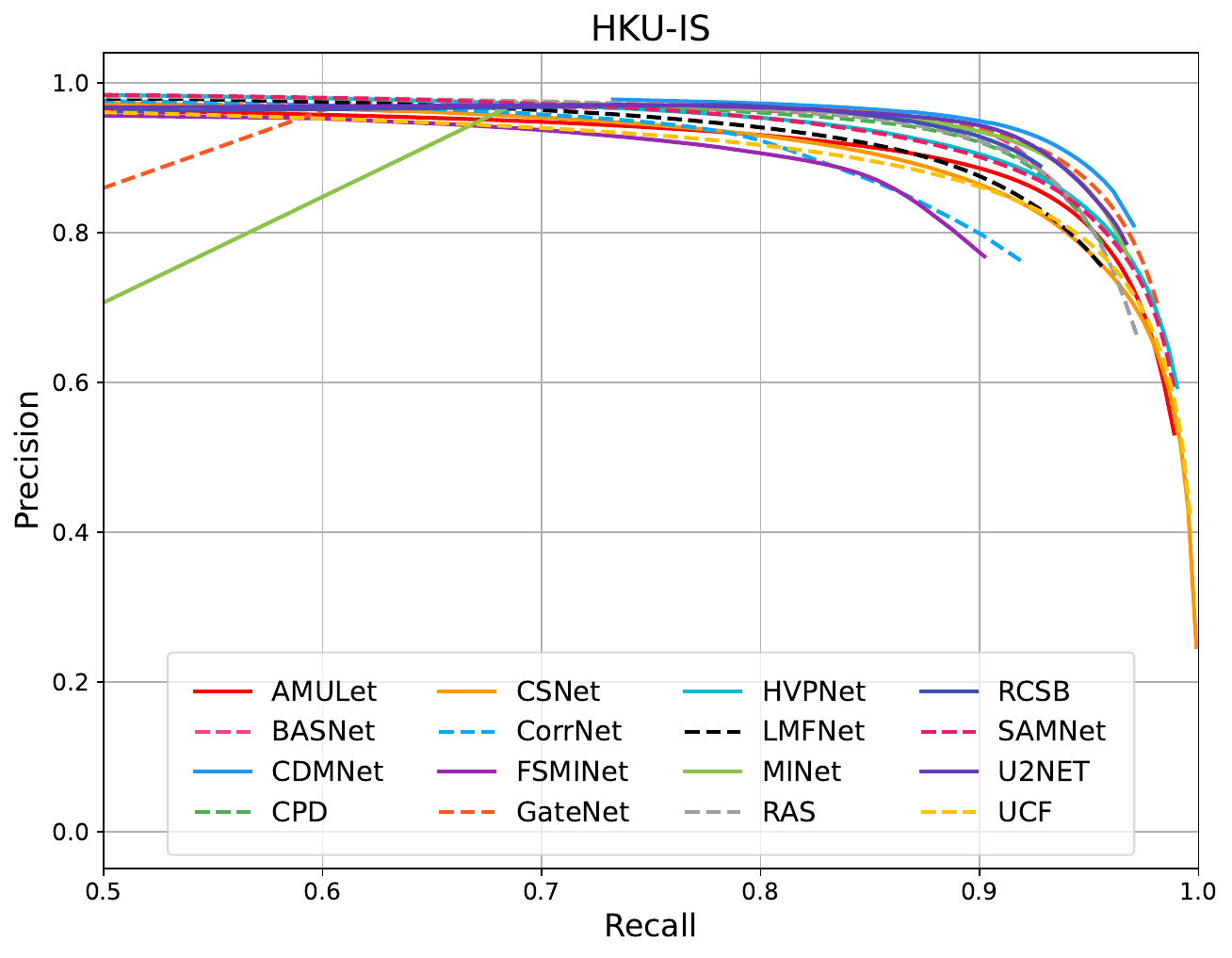} \\[1ex]
		\includegraphics[width=\linewidth]{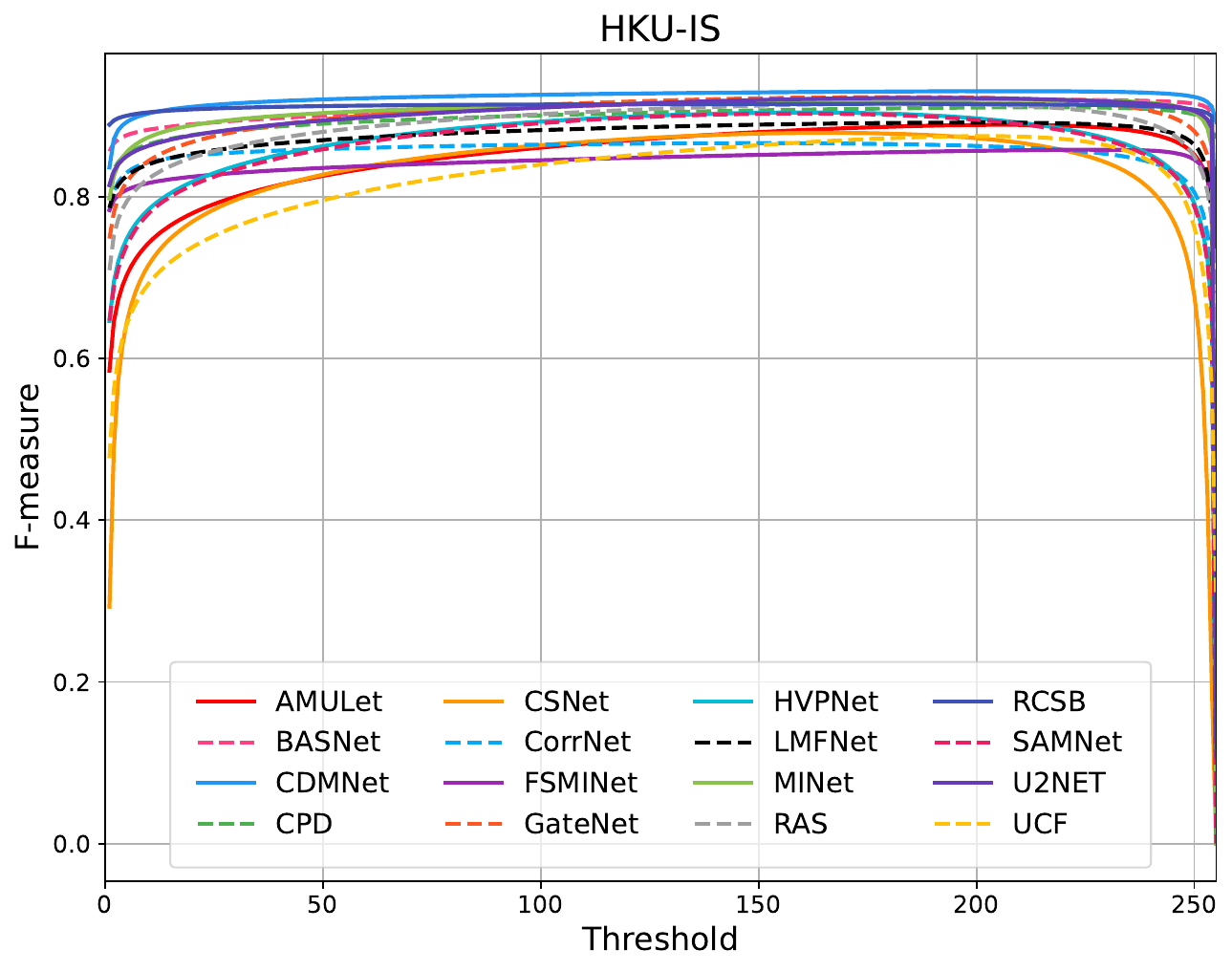} \\[1ex]
		HKU-IS
	\end{minipage}
	\hfill
	\begin{minipage}{0.162\textwidth}
		\centering
		\includegraphics[width=\linewidth]{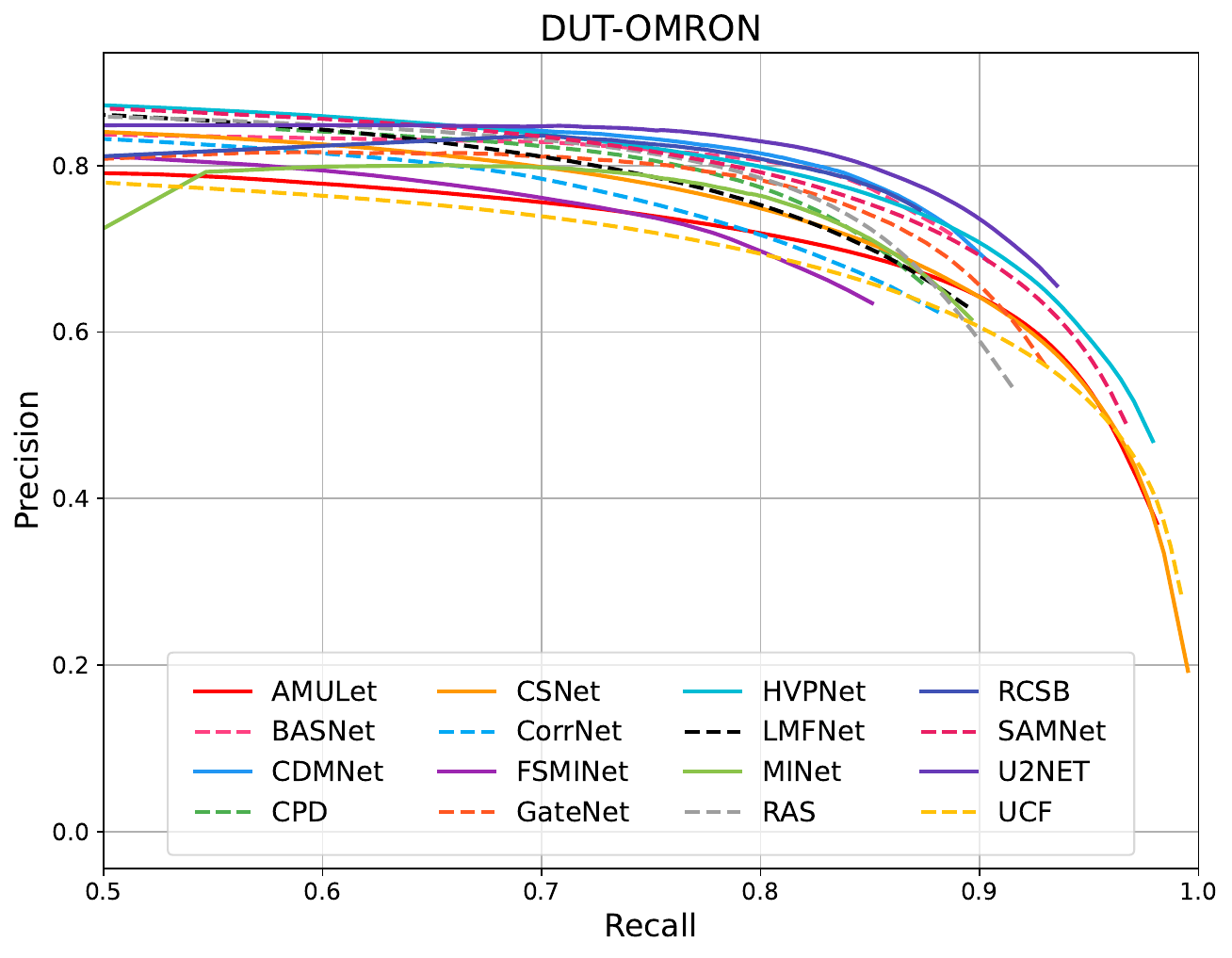} \\[1ex]
		\includegraphics[width=\linewidth]{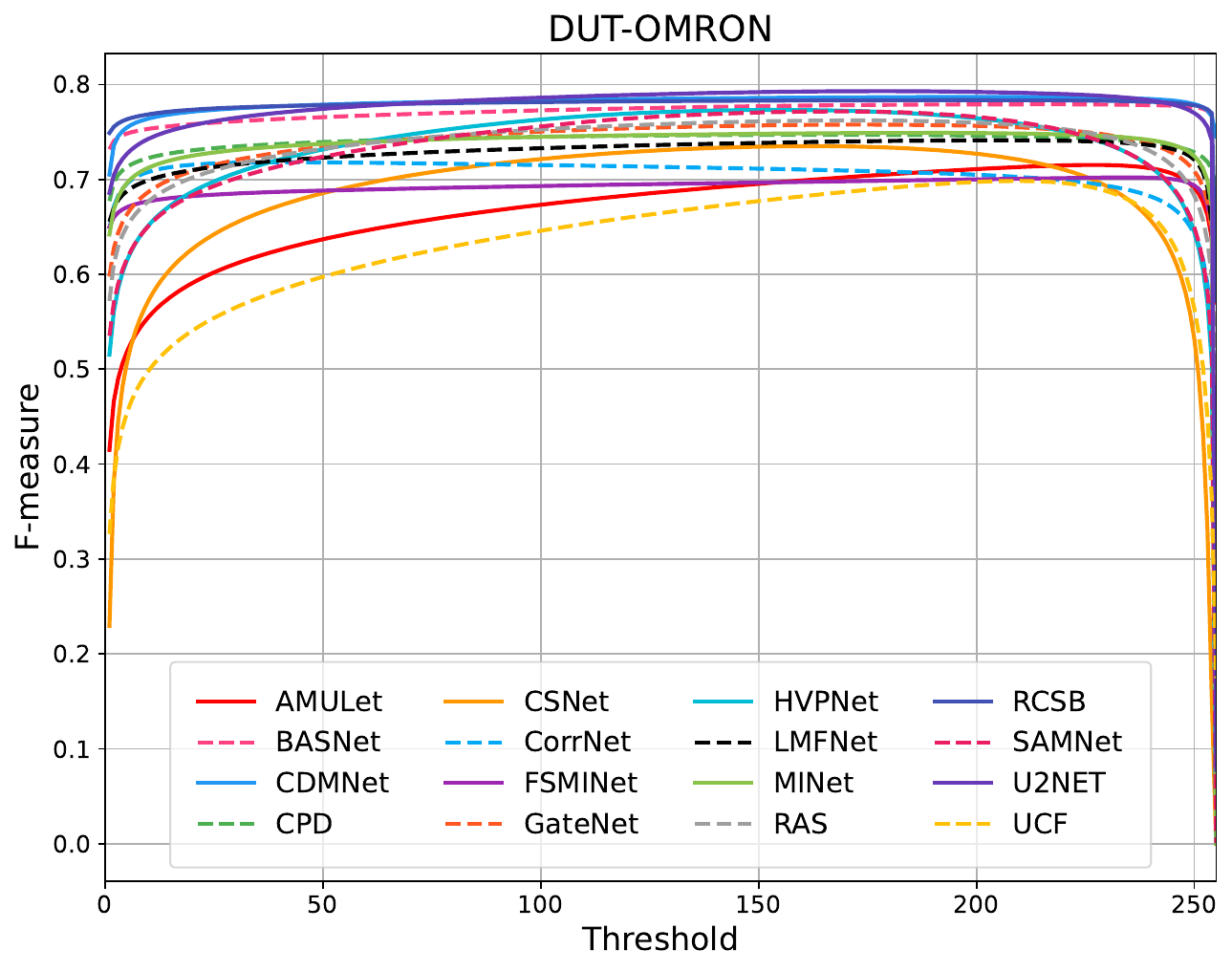} \\[1ex]
		DUT-OMRON
	\end{minipage}
	\hfill
	\begin{minipage}{0.162\textwidth}
		\centering
		\includegraphics[width=\linewidth]{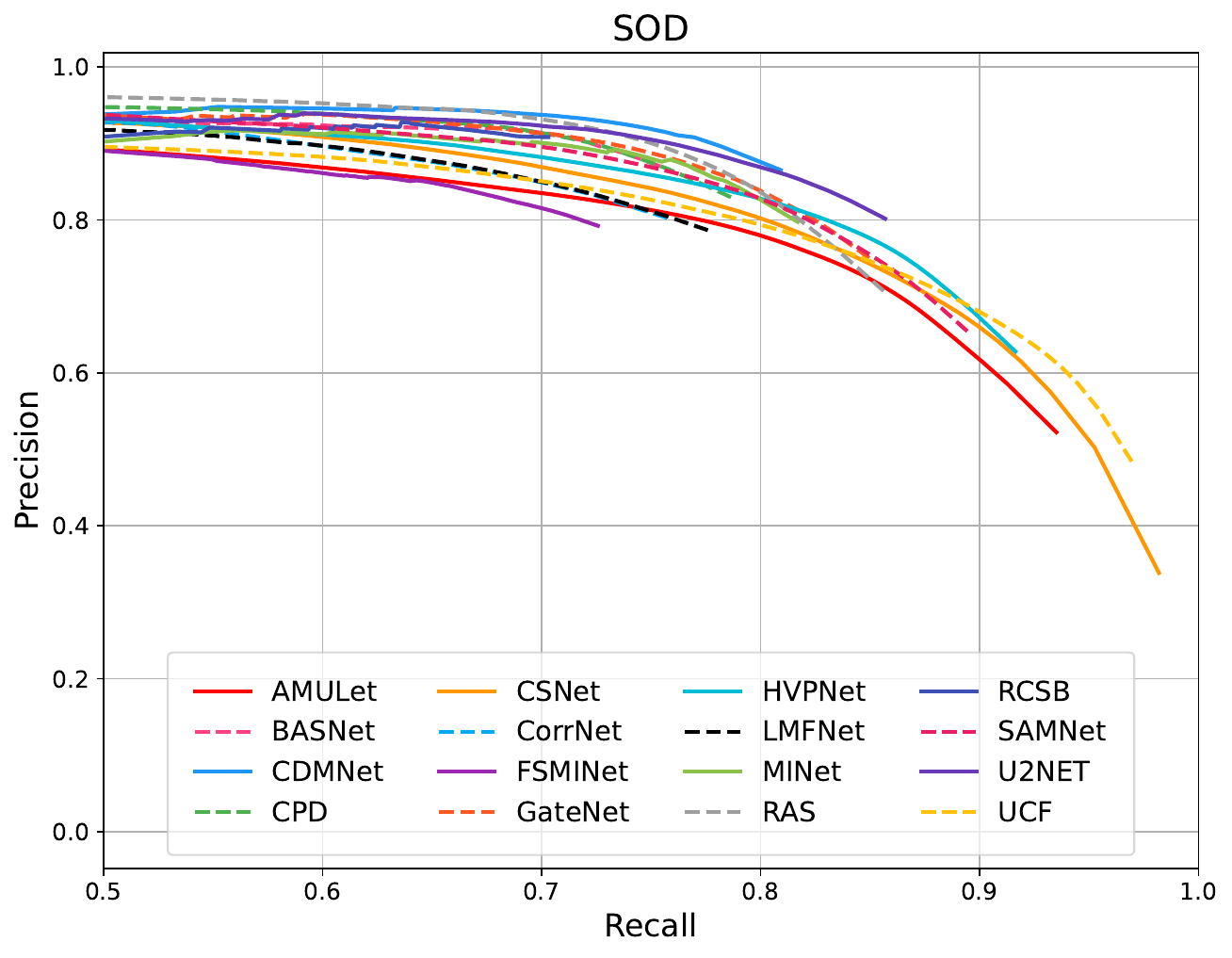} \\[1ex]
		\includegraphics[width=\linewidth]{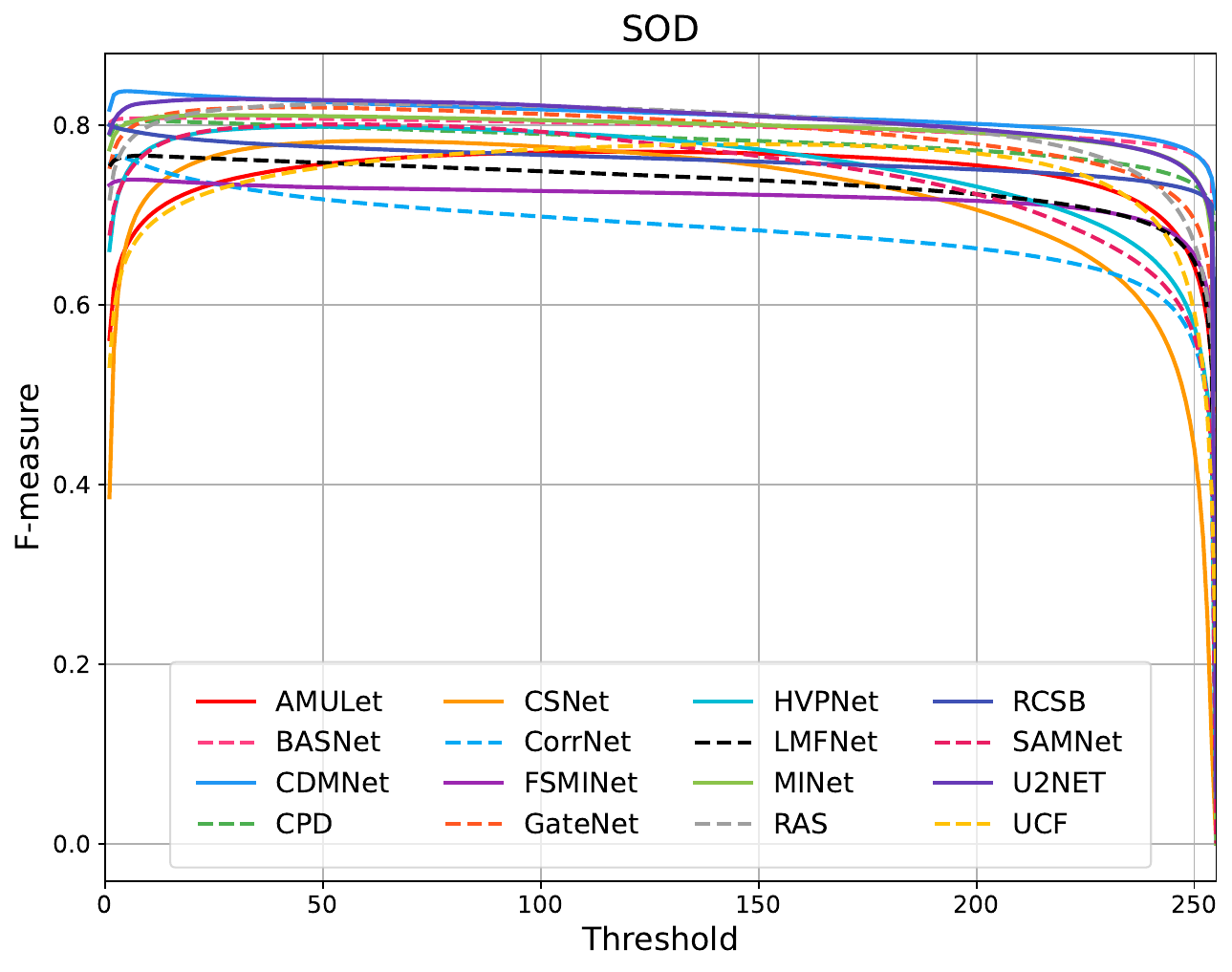} \\[1ex]
		SOD
	\end{minipage}
	\hfill
	\begin{minipage}{0.162\textwidth}
		\centering
		\includegraphics[width=\linewidth]{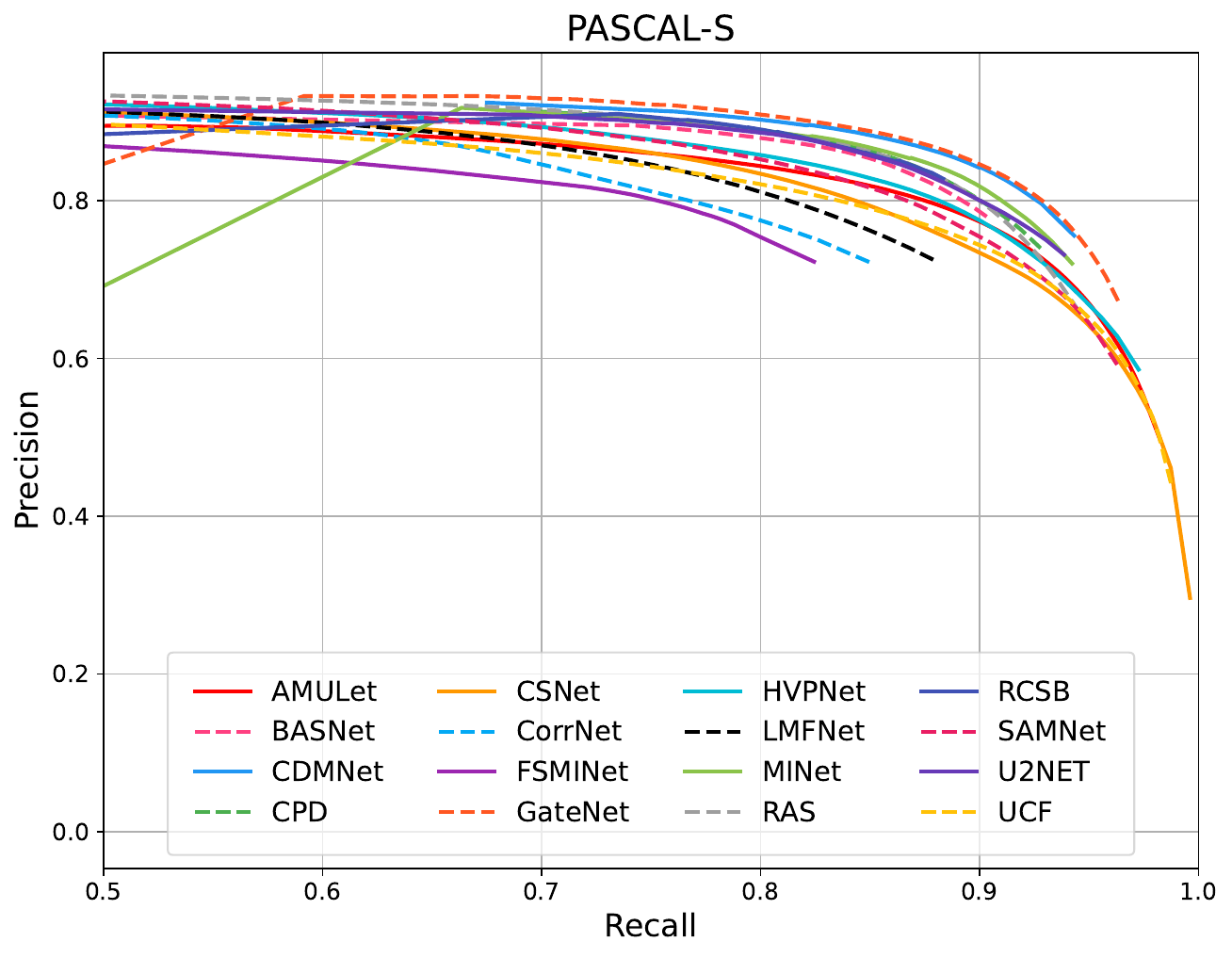} \\[1ex]
		\includegraphics[width=\linewidth]{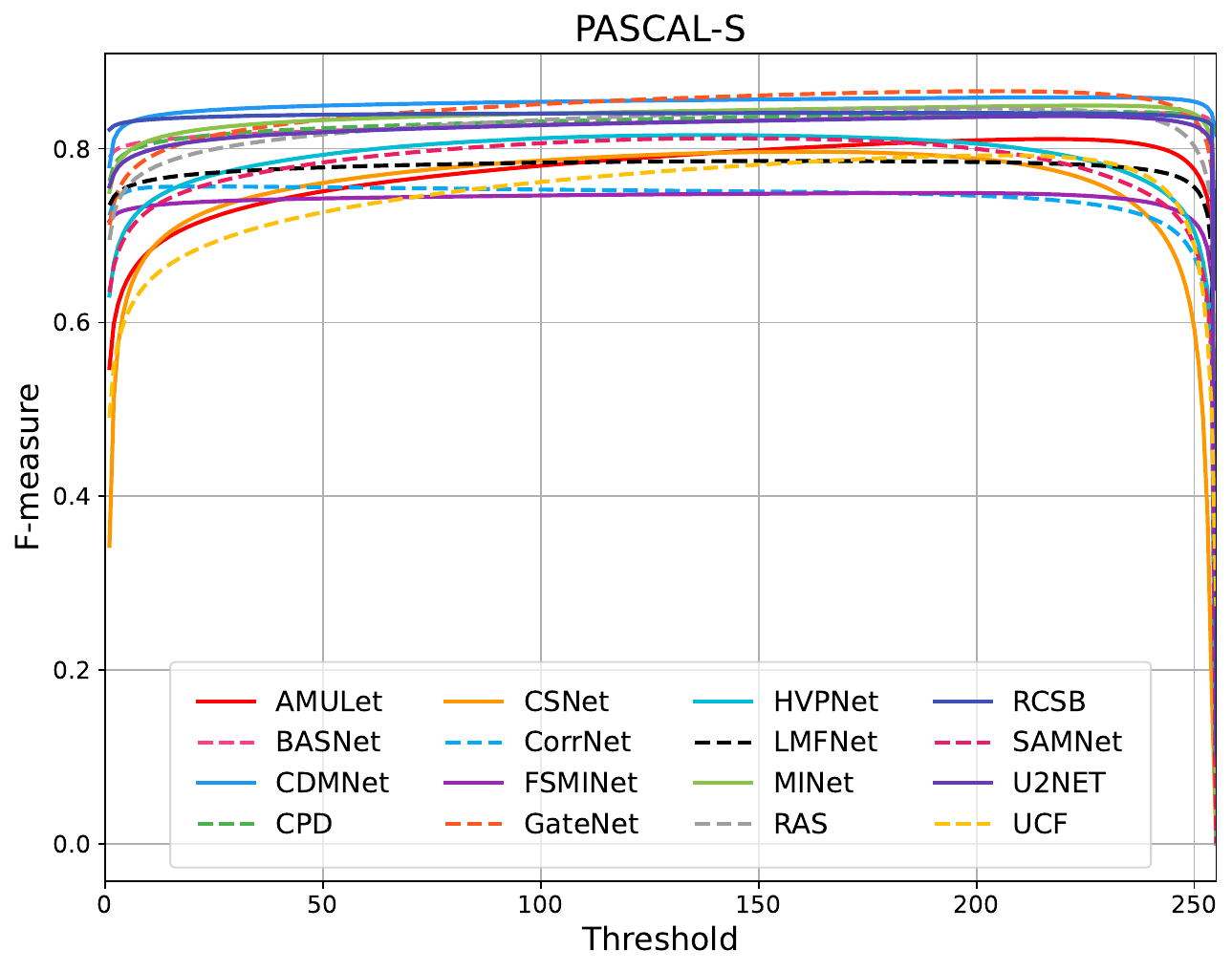} \\[1ex]
		PASCAL-S
	\end{minipage}
	
	\caption{Quantitative evaluation of different saliency object detection networks.}
	\label{fig:7}
	
	\restoregeometry 
\end{figure*}

\subsection{Performance Analysis}

We compared the proposed model with 15 state-of-the-art models, including 10 traditional models, such as AMULet\cite{zhang2017amulet}, UCF\cite{zhang2017learning}, CPD\cite{wu2019cascaded}, BASNet\cite{qin2019basnet}, GateNet\cite{zhao2020suppress}, MINet\cite{pang2020multi}, U2NET\cite{qin2020u2}, RAS\cite{chen2018reverse}, CDMNET\cite{song2023disentangle}, and RCSB\cite{ke2022recursive}, as well as 5 lightweight models, such as CSNet\cite{gao2020highly}, HVPNet\cite{liu2020lightweight}, SAMNet\cite{liu2021samnet}, FSMINet\cite{shen2022fully}, and CorrNet\cite{li2021multi}. The prediction results of all models are based on the source code provided by the authors or their publicly available results.

\textbf{Comparison with Traditional Models:} In Table \ref{tab:1}, we highlight the second-best values in gray and the best values in red for easy comparison. As shown in Fig.\ref{fig:7} and Table \ref{tab:1}, on five datasets, including ECSSD, our model performs similarly to the ten traditional models in terms of Mean Absolute Error (MAE), maximum F-measure (F$\beta$), maximum E-measure (E$\beta$), and Adaptive S-measure (Sm). Notably, on most datasets, our model outperforms UCF and AMULet. Moreover, it shows significantly fewer parameters and FLOPs compared to traditional models. For instance, compared to U2Net, which is the best performing model, our model has a slightly lower average F-measure (LMFNet: 0.820, CDMNET: 0.877) but reduces the number of parameters by 32 times and FLOPs by 5 times. As shown in Fig.\ref{fig:5}, our model achieves a good balance between performance and efficiency.

\begin{table*}[ht]
	\centering
	\resizebox{\textwidth}{!}{%
		\begin{tabular}{l|c|c|c|c|cc|cc|cc|cc|cc}
		
			\bottomrule
			\multirow{2}{*}{NO.} & \multirow{2}{*}{\textit{d}} & \multirow{2}{*}{loss} & \multirow{2}{*}{\/Param (M)} & \multirow{2}{*}{FLOPs(G)} & \multicolumn{2}{c|}{ECSSD} & \multicolumn{2}{c|}{DUTS-TE} & \multicolumn{2}{c|}{HKU-IS} &\multicolumn{2}{c|}{PASCAL-S} & \multicolumn{2}{c}{DUT-OMRON} \\
			\cmidrule{6-15}
			& & & && MAE $\downarrow$ & $F_\beta \uparrow$ & MAE $\downarrow$ & $F_\beta \uparrow$ & MAE $\downarrow$ & $F_\beta \uparrow$  & MAE $\downarrow$ & $F_\beta \uparrow$  & MAE $\downarrow$ & $F_\beta \uparrow$ \\

			\midrule
			1 & 1, 4, 12, 36, 108 & bce+ssim+iou & 0.81 & 3.8 &0.054 & 0.897  & 0.059 & 0.782 & 0.043 & 0.892 &   0.097 & 0.786 &0.062 & 0.741  \\
			2 & 1, 2, 6, 18, 54 & bce+ssim+iou & 0.81 & 3.8 & 0.056 & 0.891 & 0.060 & 0.780 & 0.044 & 0.881 & 0.096 & 0.788 & 0.067 & 0.718 \\
			3 & 1, 6, 18, 54, 162 & bce+ssim+iou & 0.81 & 3.8 & 0.059 & 0.887 & 0.061 & 0.775 & 0.045 & 0.877 & 0.102 & 0.769 & 0.063 & 0.739 \\
			4 & 1, 1, 1, 1, 1 & bce+ssim+iou  & 0.81 & 3.8 & 0.077 & 0.852 & 0.076 & 0.720 & 0.060 & 0.843 & 0.114 & 0.753 & 0.073 & 0.682 \\
			\midrule
			5 & 1, 4, 12, 36, 108 & bce & 0.81 & 3.8 & 0.064 & 0.863 & 0.064 & 0.764 & 0.052 & 0.864 & 0.101 & 0.770 & 0.070 & 0.694 \\
			
			6 & 1, 4, 12, 36, 108 & bce+ssim & 0.81 & 3.8 & 0.061 & 0.884 & 0.062 & 0.779 & 0.051 & 0.871 & 0.102 & 0.766 & 0.065 & 0.731 \\
			7 & 1, 4, 12, 36, 108 & bce+iou & 0.81 & 3.8 & 0.061 & 0.891 & 0.061 & 0.780 & 0.050 & 0.885 & 0.099 & 0.781 & 0.064 & 0.732 \\
			\midrule
			8 & 1, 4, 12, 36, 108 & bce+ssim+iou & 0.5 & 2.9 & 0.058 & 0.888 & 0.062 & 0.778 & 0.045 & 0.881 & 0.097 & 0.770 & 0.063 & 0.752 \\
			9 & 1, 4, 12, 36, 108 & bce+ssim+iou & 1.31 & 5.4 & 0.052 & 0.898 & 0.059 & 0.789 & 0.044 & 0.890 & 0.103 & 0.794 & 0.063 & 0.743 \\
			\midrule
			10 & * & bce+ssim+iou & 0.81 & 3.8 & 0.064 & 0.863 & 0.064 & 0.775 & 0.053 & 0.867 & 0.103 & 0.770 & 0.070 & 0.698 \\

			\bottomrule
		\end{tabular}
		
	}
	\caption{Quantitative comparisons between our model and different variants. }
\end{table*}

\textbf{Comparison with Lightweight Models:} In Table IV, we compare our model with existing lightweight models across five datasets. While our model performs slightly worse than HVPNet and SAMNet, it has significantly fewer parameters. Therefore, we conclude that our model is a competitive choice for devices with limited storage and computational resources.

Moreover, our model has a simpler structure compared to others, offering great application potential. It can easily be extended to other tasks, such as image classification (as demonstrated in Section VI).

In Table \ref{tab:2}, we compare the structural designs and training strategies of several models. HVPNet and SAMNet use attention mechanisms that improve performance but increase computational overhead and complicate training (e.g., requiring more data for effective learning). During training, models like CorrNet, HVPNet, and SAMNet use pretraining strategies, which enhance performance but also consume considerable computational resources and time. Therefore, our model better aligns with the application needs of lightweight networks in both its design and training process.

%

\subsection{Ablation Study}

In this section, we conduct an ablation study to evaluate some of our earlier conclusions and the model's training process across five datasets.

\textbf{The ratio of dilation rates between adjacent dilated convolution layers should be smaller than the kernel size of the previous layer.} By testing different values for the second-level dilation factor \(d[1]\), we found that changing \(d[1]\) from 2 to 4 has little impact on the model's performance, with optimal results achieved when \(d[1] = 4\).However, when \(d[1] = 6\) (with a kernel size of 5 for the first convolutional layer), model performance significantly decreases. This suggests that when the ratio of dilation rates between adjacent layers exceeds the kernel size of the previous layer, performance suffers due to information loss. Furthermore, when we set the dilation rate of each layer to 1, the model's performance dropped substantially. This demonstrates that using dilated convolutions is crucial for improving the model's performance.

\textbf{Loss Function:}  
In our model design, we used a hybrid loss function\cite{qin2019basnet} that combines SSIM, BCE, and IoU losses. To evaluate its effectiveness, we compared the model's performance with different loss functions. As shown in Table~\ref{tab:loss_comparison}, the model performed best with the hybrid loss function, validating its design.

\textbf{Model with Varying Parameters:}  
We adjusted the number of channels in the LMFNet architecture to create models with varying parameter counts (see rows 8 and 9 in Table~\ref{tab:2}). Our comparison showed that increasing the parameter count did not lead to a significant performance boost. Therefore, we conclude that the model achieves an effective balance between performance and computational cost.
﻿
In Table\ref{tab:2}-10, we removed all dilated convolutions with a dilation rate of 1, except for the first LMF layer. The results showed that performance declined due to information loss from discontinuous sampling in dilated convolutions. Thus, retaining a dilated convolution with a dilation rate of 1 in each LMFNet layer is crucial.

\section{Experiments for Extension}
Building on the success of our model in salient object detection, we conducted experiments on image classification tasks to further demonstrate the effectiveness of our network design. Specifically, in Tables\ref{tab:4} and \ref{tab:5}, \textit{OURS} refers to the encoder part of LMFNet with the intermediate and low-level feature fusion components removed, and a classifier added. \textit{OURSp} extends \textit{OURS} by increasing the number of channels to better suit the image classification task. The classifier consists of one global average pooling layer and two fully connected layers.
﻿
\subsection{Experimental Setup:}
\textbf{Dataset:} We conducted experiments on two well-known image classification datasets: CIFAR-10\cite{krizhevsky2009learning} and CIFAR-100\cite{krizhevsky2009learning}. CIFAR-10 and CIFAR-100 are classic datasets, each containing 60,000 images of 32×32 resolution, with 10 and 100 categories, respectively. The datasets are split into training and test sets in a 5:1 ratio.

\textbf{Implementation Details:} During data augmentation, we applied random cropping, flipping, and rotation to enhance model training. The experiments used the SGD optimizer with weight decay \(\lambda\). The batch size was set to 128, with an initial learning rate of 0.1, and the maximum number of epochs was set to 240. The learning rate was reduced by a factor of 0.2 at epochs 60, 120, 160, and 200. We evaluated the model on both the training and testing sets of CIFAR-10 and CIFAR-100. The results were compared with five classic lightweight backbone networks: Mobilenet~\cite{howard2017mobilenets}, Mobilenetv2\cite{sandler2018mobilenetv2}, Shufflenet\cite{zhang2018shufflenet}, Shufflenetv2\cite{ma2018shufflenet}, and Squeezenet\cite{iandola2016squeezenet}. The results were sourced from the respective codebases, ensuring identical training conditions.

﻿\begin{table}[ht]
	\centering
	\renewcommand{\arraystretch}{1.6} 
	\resizebox{0.5\textwidth}{!}{ 
		\begin{tabular}{|p{1.5cm}|p{0.8cm}|p{0.8cm}|p{0.8cm}|p{0.8cm}|p{0.8cm}|p{0.8cm}|} 
			\hline
			& \fontsize{6}{11}\selectfont CSNet& \fontsize{6}{11}\selectfont HVPNet& \fontsize{6}{11}\selectfont SAMNet & \fontsize{6}{11}\selectfont FSMINet & \fontsize{6}{11}\selectfont CorrNet & \fontsize{6}{11}\selectfont LMFNet \\
			\hline
			\fontsize{7}{11}\selectfont pre-train & \ding{55} & \checkmark & \checkmark & \ding{55} & \checkmark & \ding{55} \\
			\fontsize{7}{11}\selectfont Attention Mechanism & \ding{55} & \checkmark & \checkmark & \ding{55} & \ding{55} & \ding{55} \\
			\hline
		\end{tabular}
	}
	\caption{Comparison of different lightweight saliency object detection networks.}
	\label{tab:2}
\end{table}
	\vspace{-30pt}

\subsection{Performance Analysis:}
As shown in Table \ref{tab:4} and \ref{tab:5}, our network design achieves competitive performance on the image classification task, comparable to classical networks. Specifically, compared to lightweight backbone networks like Mobilenet and Mobilenetv2, our method uses less than half the parameters and computational resources, while delivering similar or even superior performance on datasets such as CIFAR-10 and CIFAR-100. This highlights the potential of our model design for broader applications.
﻿

﻿\begin{table}[ht]
	\raggedright
	\resizebox{0.5\textwidth}{!}{
			\begin{tabular}{lcccc}
				\hline
				\multirow{2}{*}{\textbf{Method}} & \multirow{2}{*}{\textbf{Param (M)}} & \multirow{2}{*}{\textbf{FLOPs (G)}} & \multicolumn{2}{c}{\textbf{CIFAR-100}} \\ 
				& & & \textbf{TOP1-Accuracy} & \textbf{TOP5-Accuracy} \\
				\hline
				\text{Mobilenet} & 3.32 & 0.05 & 0.6786 & 0.8886 \\
				\text{Mobilenetv2} & 2.37 & 0.07 & 0.7157 & 0.9213 \\
				\text{Shufflenet} & 1.01 & 0.05 & 0.7178 & 0.9143 \\
				\text{Squeezenet} & 0.78 & 0.06  & 0.705  & 0.9141 \\
				\text{Shufflenetv2} & 1.36 & 0.05 & 0.7179 & 0.9169 \\
				\text{OURS} & 0.66 & 0.04 & 0.6995 & 0.9130 \\
				\text{OURSp} & 1.34 & 0.1 & 0.7135 & 0.9134 \\
				\hline
			\end{tabular}
		}
		
		\caption{Comparison of Models on CIFAR-100}
		\label{tab:4}
	\end{table}
	\vspace{-40pt} 
	\begin{table}[ht]
		\raggedright
		\resizebox{0.5\textwidth}{!}{
			\begin{tabular}{lccccc}
				\hline
				\multirow{2}{*}{\textbf{Method}} & \multirow{2}{*}{\textbf{Param (M)}} & \multirow{2}{*}{\textbf{FLOPs}} & \multicolumn{2}{c}{\textbf{CIFAR-10}} \\ 
				& & & \textbf{TOP1-Accuracy} & \textbf{TOP5-Accuracy} \\
				\hline
				\text{Mobilenet} & 3.22 & 0.05 & 0.9210 & 0.9975 \\
				\text{Mobilenetv2} & 2.25 &0.07 & 0.9211 & 0.9975 \\
				\text{Shufflenet} & 0.93 & 0.05  & 0.9237 & 0.9973 \\
				\text{Squeezenet} & 0.73 & 0.05  & 0.9279 & 0.9980 \\
				\text{Shufflenetv2} & 1.27 & 0.05 & 0.9290 & 0.9966 \\
				\text{OURS} & 0.62& 0.04 & 0.9155 & 0.9974 \\
				\text{OURSp} & 1.31 & 0.1 & 0.9234 & 0.9978 \\
				\hline
			\end{tabular}
		}
		
		\caption{Comparison of Models on CIFAR-10}
		\label{tab:5}
	\end{table}
	\vspace{-40pt} 
\section{CONCLUSION}
To address the challenge of extracting and utilizing multi-scale information in lightweight networks, this paper proposes a \textbf{Lightweight Multi-scale Feature (LMF) layer}, which employs a fully connected structure based on the receptive field of neural networks. In the LMF layer, depthwise separable dilated convolutions are integrated through a fully connected structure to effectively capture and utilize multi-scale information.  

Building on this, we developed \textbf{LMFNet}, a lightweight network tailored for salient object detection. LMFNet employs multiple stacked LMF layers, allowing for diverse receptive fields and enabling comprehensive multi-scale feature extraction. Extensive experiments on five benchmark datasets—DUTS-TE, ECSSD, HKU-IS, PASCAL-S, and DUT-OMRON—demonstrate that LMFNet achieves competitive or superior detection performance compared to state-of-the-art saliency models, while requiring only \textbf{0.81M} parameters. Additionally, supplementary experiments on \textbf{CIFAR-10} and \textbf{CIFAR-100} suggest that our model design holds potential for broader applications.  

Experimental results further indicate that increasing the parameter count does not significantly enhance model performance, suggesting redundancy in the design. Thus, in future work, we plan to explore model pruning and other optimization techniques to further improve efficiency and reduce computational cost.

\setlength{\bibsep}{6pt}
\bibliography{reference}


\begin{thebibliography}{49}
\ifx \bisbn   \undefined \def \bisbn  #1{ISBN #1}\fi
\ifx \binits  \undefined \def \binits#1{#1}\fi
\ifx \bauthor  \undefined \def \bauthor#1{#1}\fi
\ifx \batitle  \undefined \def \batitle#1{#1}\fi
\ifx \bjtitle  \undefined \def \bjtitle#1{#1}\fi
\ifx \bvolume  \undefined \def \bvolume#1{\textbf{#1}}\fi
\ifx \byear  \undefined \def \byear#1{#1}\fi
\ifx \bissue  \undefined \def \bissue#1{#1}\fi
\ifx \bfpage  \undefined \def \bfpage#1{#1}\fi
\ifx \blpage  \undefined \def \blpage #1{#1}\fi
\ifx \burl  \undefined \def \burl#1{\textsf{#1}}\fi
\ifx \doiurl  \undefined \def \doiurl#1{\url{https://doi.org/#1}}\fi
\ifx \betal  \undefined \def \betal{\textit{et al.}}\fi
\ifx \binstitute  \undefined \def \binstitute#1{#1}\fi
\ifx \binstitutionaled  \undefined \def \binstitutionaled#1{#1}\fi
\ifx \bctitle  \undefined \def \bctitle#1{#1}\fi
\ifx \beditor  \undefined \def \beditor#1{#1}\fi
\ifx \bpublisher  \undefined \def \bpublisher#1{#1}\fi
\ifx \bbtitle  \undefined \def \bbtitle#1{#1}\fi
\ifx \bedition  \undefined \def \bedition#1{#1}\fi
\ifx \bseriesno  \undefined \def \bseriesno#1{#1}\fi
\ifx \blocation  \undefined \def \blocation#1{#1}\fi
\ifx \bsertitle  \undefined \def \bsertitle#1{#1}\fi
\ifx \bsnm \undefined \def \bsnm#1{#1}\fi
\ifx \bsuffix \undefined \def \bsuffix#1{#1}\fi
\ifx \bparticle \undefined \def \bparticle#1{#1}\fi
\ifx \barticle \undefined \def \barticle#1{#1}\fi
\bibcommenthead
\ifx \bconfdate \undefined \def \bconfdate #1{#1}\fi
\ifx \botherref \undefined \def \botherref #1{#1}\fi
\ifx \url \undefined \def \url#1{\textsf{#1}}\fi
\ifx \bchapter \undefined \def \bchapter#1{#1}\fi
\ifx \bbook \undefined \def \bbook#1{#1}\fi
\ifx \bcomment \undefined \def \bcomment#1{#1}\fi
\ifx \oauthor \undefined \def \oauthor#1{#1}\fi
\ifx \citeauthoryear \undefined \def \citeauthoryear#1{#1}\fi
\ifx \endbibitem  \undefined \def \endbibitem {}\fi
\ifx \bconflocation  \undefined \def \bconflocation#1{#1}\fi
\ifx \arxivurl  \undefined \def \arxivurl#1{\textsf{#1}}\fi
\csname PreBibitemsHook\endcsname

\bibitem[\protect\citeauthoryear{Krizhevsky
  et~al.}{2012}]{krizhevsky2012imagenet}
\begin{botherref}
\oauthor{\bsnm{Krizhevsky}, \binits{A.}},
\oauthor{\bsnm{Sutskever}, \binits{I.}},
\oauthor{\bsnm{Hinton}, \binits{G.E.}}:
Imagenet classification with deep convolutional neural networks.
Advances in neural information processing systems
\textbf{25}
(2012)
\end{botherref}
\endbibitem

\bibitem[\protect\citeauthoryear{Simonyan and
  Zisserman}{2014}]{simonyan2014very}
\begin{botherref}
\oauthor{\bsnm{Simonyan}, \binits{K.}},
\oauthor{\bsnm{Zisserman}, \binits{A.}}:
Very deep convolutional networks for large-scale image recognition.
arXiv preprint arXiv:1409.1556
(2014)
\end{botherref}
\endbibitem

\bibitem[\protect\citeauthoryear{He et~al.}{2016}]{he2016deep}
\begin{bchapter}
\bauthor{\bsnm{He}, \binits{K.}},
\bauthor{\bsnm{Zhang}, \binits{X.}},
\bauthor{\bsnm{Ren}, \binits{S.}},
\bauthor{\bsnm{Sun}, \binits{J.}}:
\bctitle{Deep residual learning for image recognition}.
In: \bbtitle{Proceedings of the IEEE Conference on Computer Vision and Pattern
  Recognition},
pp. \bfpage{770}--\blpage{778}
(\byear{2016})
\end{bchapter}
\endbibitem

\bibitem[\protect\citeauthoryear{Szegedy et~al.}{2015}]{szegedy2015going}
\begin{bchapter}
\bauthor{\bsnm{Szegedy}, \binits{C.}},
\bauthor{\bsnm{Liu}, \binits{W.}},
\bauthor{\bsnm{Jia}, \binits{Y.}},
\bauthor{\bsnm{Sermanet}, \binits{P.}},
\bauthor{\bsnm{Reed}, \binits{S.}},
\bauthor{\bsnm{Anguelov}, \binits{D.}},
\bauthor{\bsnm{Erhan}, \binits{D.}},
\bauthor{\bsnm{Vanhoucke}, \binits{V.}},
\bauthor{\bsnm{Rabinovich}, \binits{A.}}:
\bctitle{Going deeper with convolutions}.
In: \bbtitle{Proceedings of the IEEE Conference on Computer Vision and Pattern
  Recognition},
pp. \bfpage{1}--\blpage{9}
(\byear{2015})
\end{bchapter}
\endbibitem

\bibitem[\protect\citeauthoryear{Lin et~al.}{2017}]{lin2017feature}
\begin{bchapter}
\bauthor{\bsnm{Lin}, \binits{T.-Y.}},
\bauthor{\bsnm{Doll{\'a}r}, \binits{P.}},
\bauthor{\bsnm{Girshick}, \binits{R.}},
\bauthor{\bsnm{He}, \binits{K.}},
\bauthor{\bsnm{Hariharan}, \binits{B.}},
\bauthor{\bsnm{Belongie}, \binits{S.}}:
\bctitle{Feature pyramid networks for object detection}.
In: \bbtitle{Proceedings of the IEEE Conference on Computer Vision and Pattern
  Recognition},
pp. \bfpage{2117}--\blpage{2125}
(\byear{2017})
\end{bchapter}
\endbibitem

\bibitem[\protect\citeauthoryear{Zhou et~al.}{2023}]{zhou2023wavenet}
\begin{barticle}
\bauthor{\bsnm{Zhou}, \binits{W.}},
\bauthor{\bsnm{Sun}, \binits{F.}},
\bauthor{\bsnm{Jiang}, \binits{Q.}},
\bauthor{\bsnm{Cong}, \binits{R.}},
\bauthor{\bsnm{Hwang}, \binits{J.-N.}}:
\batitle{Wavenet: Wavelet network with knowledge distillation for rgb-t salient
  object detection}.
\bjtitle{IEEE Transactions on Image Processing}
\bvolume{32},
\bfpage{3027}--\blpage{3039}
(\byear{2023})
\end{barticle}
\endbibitem

\bibitem[\protect\citeauthoryear{Liu et~al.}{2017}]{liu2017learning}
\begin{bchapter}
\bauthor{\bsnm{Liu}, \binits{Z.}},
\bauthor{\bsnm{Li}, \binits{J.}},
\bauthor{\bsnm{Shen}, \binits{Z.}},
\bauthor{\bsnm{Huang}, \binits{G.}},
\bauthor{\bsnm{Yan}, \binits{S.}},
\bauthor{\bsnm{Zhang}, \binits{C.}}:
\bctitle{Learning efficient convolutional networks through network slimming}.
In: \bbtitle{Proceedings of the IEEE International Conference on Computer
  Vision},
pp. \bfpage{2736}--\blpage{2744}
(\byear{2017})
\end{bchapter}
\endbibitem

\bibitem[\protect\citeauthoryear{Zhou et~al.}{2023}]{zhou2023lsnet}
\begin{barticle}
\bauthor{\bsnm{Zhou}, \binits{W.}},
\bauthor{\bsnm{Zhu}, \binits{Y.}},
\bauthor{\bsnm{Lei}, \binits{J.}},
\bauthor{\bsnm{Yang}, \binits{R.}},
\bauthor{\bsnm{Yu}, \binits{L.}}:
\batitle{Lsnet: Lightweight spatial boosting network for detecting salient
  objects in rgb-thermal images}.
\bjtitle{IEEE Transactions on Image Processing}
\bvolume{32},
\bfpage{1329}--\blpage{1340}
(\byear{2023})
\end{barticle}
\endbibitem

\bibitem[\protect\citeauthoryear{Liu et~al.}{2020}]{liu2020lightweight}
\begin{barticle}
\bauthor{\bsnm{Liu}, \binits{Y.}},
\bauthor{\bsnm{Gu}, \binits{Y.-C.}},
\bauthor{\bsnm{Zhang}, \binits{X.-Y.}},
\bauthor{\bsnm{Wang}, \binits{W.}},
\bauthor{\bsnm{Cheng}, \binits{M.-M.}}:
\batitle{Lightweight salient object detection via hierarchical visual
  perception learning}.
\bjtitle{IEEE transactions on cybernetics}
\bvolume{51}(\bissue{9}),
\bfpage{4439}--\blpage{4449}
(\byear{2020})
\end{barticle}
\endbibitem

\bibitem[\protect\citeauthoryear{Jiang et~al.}{2013}]{jiang2013salient}
\begin{bchapter}
\bauthor{\bsnm{Jiang}, \binits{H.}},
\bauthor{\bsnm{Wang}, \binits{J.}},
\bauthor{\bsnm{Yuan}, \binits{Z.}},
\bauthor{\bsnm{Wu}, \binits{Y.}},
\bauthor{\bsnm{Zheng}, \binits{N.}},
\bauthor{\bsnm{Li}, \binits{S.}}:
\bctitle{Salient object detection: A discriminative regional feature
  integration approach}.
In: \bbtitle{Proceedings of the IEEE Conference on Computer Vision and Pattern
  Recognition},
pp. \bfpage{2083}--\blpage{2090}
(\byear{2013})
\end{bchapter}
\endbibitem

\bibitem[\protect\citeauthoryear{Li et~al.}{2013}]{li2013saliency}
\begin{bchapter}
\bauthor{\bsnm{Li}, \binits{X.}},
\bauthor{\bsnm{Lu}, \binits{H.}},
\bauthor{\bsnm{Zhang}, \binits{L.}},
\bauthor{\bsnm{Ruan}, \binits{X.}},
\bauthor{\bsnm{Yang}, \binits{M.-H.}}:
\bctitle{Saliency detection via dense and sparse reconstruction}.
In: \bbtitle{Proceedings of the IEEE International Conference on Computer
  Vision},
pp. \bfpage{2976}--\blpage{2983}
(\byear{2013})
\end{bchapter}
\endbibitem

\bibitem[\protect\citeauthoryear{Nawaz and Yan}{2020}]{nawaz2020saliency}
\begin{barticle}
\bauthor{\bsnm{Nawaz}, \binits{M.}},
\bauthor{\bsnm{Yan}, \binits{H.}}:
\batitle{Saliency detection using deep features and affinity-based robust
  background subtraction}.
\bjtitle{IEEE transactions on multimedia}
\bvolume{23},
\bfpage{2902}--\blpage{2916}
(\byear{2020})
\end{barticle}
\endbibitem

\bibitem[\protect\citeauthoryear{Liu et~al.}{2018}]{liu2018picanet}
\begin{bchapter}
\bauthor{\bsnm{Liu}, \binits{N.}},
\bauthor{\bsnm{Han}, \binits{J.}},
\bauthor{\bsnm{Yang}, \binits{M.-H.}}:
\bctitle{Picanet: Learning pixel-wise contextual attention for saliency
  detection}.
In: \bbtitle{Proceedings of the IEEE Conference on Computer Vision and Pattern
  Recognition},
pp. \bfpage{3089}--\blpage{3098}
(\byear{2018})
\end{bchapter}
\endbibitem

\bibitem[\protect\citeauthoryear{Qin et~al.}{2020}]{qin2020u2}
\begin{barticle}
\bauthor{\bsnm{Qin}, \binits{X.}},
\bauthor{\bsnm{Zhang}, \binits{Z.}},
\bauthor{\bsnm{Huang}, \binits{C.}},
\bauthor{\bsnm{Dehghan}, \binits{M.}},
\bauthor{\bsnm{Zaiane}, \binits{O.R.}},
\bauthor{\bsnm{Jagersand}, \binits{M.}}:
\batitle{U2-net: Going deeper with nested u-structure for salient object
  detection}.
\bjtitle{Pattern recognition}
\bvolume{106},
\bfpage{107404}
(\byear{2020})
\end{barticle}
\endbibitem

\bibitem[\protect\citeauthoryear{Ke and Tsubono}{2022}]{ke2022recursive}
\begin{bchapter}
\bauthor{\bsnm{Ke}, \binits{Y.Y.}},
\bauthor{\bsnm{Tsubono}, \binits{T.}}:
\bctitle{Recursive contour-saliency blending network for accurate salient
  object detection}.
In: \bbtitle{Proceedings of the IEEE/CVF Winter Conference on Applications of
  Computer Vision},
pp. \bfpage{2940}--\blpage{2950}
(\byear{2022})
\end{bchapter}
\endbibitem

\bibitem[\protect\citeauthoryear{Pang et~al.}{2020}]{pang2020multi}
\begin{bchapter}
\bauthor{\bsnm{Pang}, \binits{Y.}},
\bauthor{\bsnm{Zhao}, \binits{X.}},
\bauthor{\bsnm{Zhang}, \binits{L.}},
\bauthor{\bsnm{Lu}, \binits{H.}}:
\bctitle{Multi-scale interactive network for salient object detection}.
In: \bbtitle{Proceedings of the IEEE/CVF Conference on Computer Vision and
  Pattern Recognition},
pp. \bfpage{9413}--\blpage{9422}
(\byear{2020})
\end{bchapter}
\endbibitem

\bibitem[\protect\citeauthoryear{Liu et~al.}{2021}]{liu2021samnet}
\begin{barticle}
\bauthor{\bsnm{Liu}, \binits{Y.}},
\bauthor{\bsnm{Zhang}, \binits{X.-Y.}},
\bauthor{\bsnm{Bian}, \binits{J.-W.}},
\bauthor{\bsnm{Zhang}, \binits{L.}},
\bauthor{\bsnm{Cheng}, \binits{M.-M.}}:
\batitle{Samnet: Stereoscopically attentive multi-scale network for lightweight
  salient object detection}.
\bjtitle{IEEE Transactions on Image Processing}
\bvolume{30},
\bfpage{3804}--\blpage{3814}
(\byear{2021})
\end{barticle}
\endbibitem

\bibitem[\protect\citeauthoryear{Zhou et~al.}{2024}]{zhou2024admnet}
\begin{botherref}
\oauthor{\bsnm{Zhou}, \binits{X.}},
\oauthor{\bsnm{Shen}, \binits{K.}},
\oauthor{\bsnm{Liu}, \binits{Z.}}:
Admnet: Attention-guided densely multi-scale network for lightweight salient
  object detection.
IEEE Transactions on Multimedia
(2024)
\end{botherref}
\endbibitem

\bibitem[\protect\citeauthoryear{Liu et~al.}{2010}]{liu2010learning}
\begin{barticle}
\bauthor{\bsnm{Liu}, \binits{T.}},
\bauthor{\bsnm{Yuan}, \binits{Z.}},
\bauthor{\bsnm{Sun}, \binits{J.}},
\bauthor{\bsnm{Wang}, \binits{J.}},
\bauthor{\bsnm{Zheng}, \binits{N.}},
\bauthor{\bsnm{Tang}, \binits{X.}},
\bauthor{\bsnm{Shum}, \binits{H.-Y.}}:
\batitle{Learning to detect a salient object}.
\bjtitle{IEEE Transactions on Pattern analysis and machine intelligence}
\bvolume{33}(\bissue{2}),
\bfpage{353}--\blpage{367}
(\byear{2010})
\end{barticle}
\endbibitem

\bibitem[\protect\citeauthoryear{Achanta et~al.}{2009}]{achanta2009frequency}
\begin{bchapter}
\bauthor{\bsnm{Achanta}, \binits{R.}},
\bauthor{\bsnm{Hemami}, \binits{S.}},
\bauthor{\bsnm{Estrada}, \binits{F.}},
\bauthor{\bsnm{Susstrunk}, \binits{S.}}:
\bctitle{Frequency-tuned salient region detection}.
In: \bbtitle{2009 IEEE Conference on Computer Vision and Pattern Recognition},
pp. \bfpage{1597}--\blpage{1604}
(\byear{2009}).
\bcomment{IEEE}
\end{bchapter}
\endbibitem

\bibitem[\protect\citeauthoryear{Kuo}{2016}]{kuo2016understanding}
\begin{barticle}
\bauthor{\bsnm{Kuo}, \binits{C.-C.J.}}:
\batitle{Understanding convolutional neural networks with a mathematical
  model}.
\bjtitle{Journal of Visual Communication and Image Representation}
\bvolume{41},
\bfpage{406}--\blpage{413}
(\byear{2016})
\end{barticle}
\endbibitem

\bibitem[\protect\citeauthoryear{Qiu et~al.}{2022}]{qiu2022a2sppnet}
\begin{barticle}
\bauthor{\bsnm{Qiu}, \binits{Y.}},
\bauthor{\bsnm{Liu}, \binits{Y.}},
\bauthor{\bsnm{Chen}, \binits{Y.}},
\bauthor{\bsnm{Zhang}, \binits{J.}},
\bauthor{\bsnm{Zhu}, \binits{J.}},
\bauthor{\bsnm{Xu}, \binits{J.}}:
\batitle{A2sppnet: Attentive atrous spatial pyramid pooling network for salient
  object detection}.
\bjtitle{IEEE Transactions on Multimedia}
\bvolume{25},
\bfpage{1991}--\blpage{2006}
(\byear{2022})
\end{barticle}
\endbibitem

\bibitem[\protect\citeauthoryear{Howard et~al.}{2017}]{howard2017mobilenets}
\begin{botherref}
\oauthor{\bsnm{Howard}, \binits{A.G.}},
\oauthor{\bsnm{Zhu}, \binits{M.}},
\oauthor{\bsnm{Chen}, \binits{B.}},
\oauthor{\bsnm{Kalenichenko}, \binits{D.}},
\oauthor{\bsnm{Wang}, \binits{W.}},
\oauthor{\bsnm{Weyand}, \binits{T.}},
\oauthor{\bsnm{Andreetto}, \binits{M.}},
\oauthor{\bsnm{Adam}, \binits{H.}}:
Mobilenets: Efficient convolutional neural networks for mobile vision
  applications.
arXiv preprint arXiv:1704.04861
(2017)
\end{botherref}
\endbibitem

\bibitem[\protect\citeauthoryear{Chen et~al.}{2017}]{chen2017deeplab}
\begin{barticle}
\bauthor{\bsnm{Chen}, \binits{L.-C.}},
\bauthor{\bsnm{Papandreou}, \binits{G.}},
\bauthor{\bsnm{Kokkinos}, \binits{I.}},
\bauthor{\bsnm{Murphy}, \binits{K.}},
\bauthor{\bsnm{Yuille}, \binits{A.L.}}:
\batitle{Deeplab: Semantic image segmentation with deep convolutional nets,
  atrous convolution, and fully connected crfs}.
\bjtitle{IEEE transactions on pattern analysis and machine intelligence}
\bvolume{40}(\bissue{4}),
\bfpage{834}--\blpage{848}
(\byear{2017})
\end{barticle}
\endbibitem

\bibitem[\protect\citeauthoryear{Wang et~al.}{2003}]{wang2003multiscale}
\begin{bchapter}
\bauthor{\bsnm{Wang}, \binits{Z.}},
\bauthor{\bsnm{Simoncelli}, \binits{E.P.}},
\bauthor{\bsnm{Bovik}, \binits{A.C.}}:
\bctitle{Multiscale structural similarity for image quality assessment}.
In: \bbtitle{The Thrity-Seventh Asilomar Conference on Signals, Systems \&
  Computers, 2003},
vol. \bseriesno{2},
pp. \bfpage{1398}--\blpage{1402}
(\byear{2003}).
\bcomment{Ieee}
\end{bchapter}
\endbibitem

\bibitem[\protect\citeauthoryear{De~Boer et~al.}{2005}]{de2005tutorial}
\begin{barticle}
\bauthor{\bsnm{De~Boer}, \binits{P.-T.}},
\bauthor{\bsnm{Kroese}, \binits{D.P.}},
\bauthor{\bsnm{Mannor}, \binits{S.}},
\bauthor{\bsnm{Rubinstein}, \binits{R.Y.}}:
\batitle{A tutorial on the cross-entropy method}.
\bjtitle{Annals of operations research}
\bvolume{134},
\bfpage{19}--\blpage{67}
(\byear{2005})
\end{barticle}
\endbibitem

\bibitem[\protect\citeauthoryear{M{\'a}ttyus
  et~al.}{2017}]{mattyus2017deeproadmapper}
\begin{bchapter}
\bauthor{\bsnm{M{\'a}ttyus}, \binits{G.}},
\bauthor{\bsnm{Luo}, \binits{W.}},
\bauthor{\bsnm{Urtasun}, \binits{R.}}:
\bctitle{Deeproadmapper: Extracting road topology from aerial images}.
In: \bbtitle{Proceedings of the IEEE International Conference on Computer
  Vision},
pp. \bfpage{3438}--\blpage{3446}
(\byear{2017})
\end{bchapter}
\endbibitem

\bibitem[\protect\citeauthoryear{Wang et~al.}{2017}]{wang2017learning}
\begin{bchapter}
\bauthor{\bsnm{Wang}, \binits{L.}},
\bauthor{\bsnm{Lu}, \binits{H.}},
\bauthor{\bsnm{Wang}, \binits{Y.}},
\bauthor{\bsnm{Feng}, \binits{M.}},
\bauthor{\bsnm{Wang}, \binits{D.}},
\bauthor{\bsnm{Yin}, \binits{B.}},
\bauthor{\bsnm{Ruan}, \binits{X.}}:
\bctitle{Learning to detect salient objects with image-level supervision}.
In: \bbtitle{Proceedings of the IEEE Conference on Computer Vision and Pattern
  Recognition},
pp. \bfpage{136}--\blpage{145}
(\byear{2017})
\end{bchapter}
\endbibitem

\bibitem[\protect\citeauthoryear{Yan et~al.}{2013}]{yan2013hierarchical}
\begin{bchapter}
\bauthor{\bsnm{Yan}, \binits{Q.}},
\bauthor{\bsnm{Xu}, \binits{L.}},
\bauthor{\bsnm{Shi}, \binits{J.}},
\bauthor{\bsnm{Jia}, \binits{J.}}:
\bctitle{Hierarchical saliency detection}.
In: \bbtitle{Proceedings of the IEEE Conference on Computer Vision and Pattern
  Recognition},
pp. \bfpage{1155}--\blpage{1162}
(\byear{2013})
\end{bchapter}
\endbibitem

\bibitem[\protect\citeauthoryear{Li and Yu}{2015}]{li2015visual}
\begin{bchapter}
\bauthor{\bsnm{Li}, \binits{G.}},
\bauthor{\bsnm{Yu}, \binits{Y.}}:
\bctitle{Visual saliency based on multiscale deep features}.
In: \bbtitle{Proceedings of the IEEE Conference on Computer Vision and Pattern
  Recognition},
pp. \bfpage{5455}--\blpage{5463}
(\byear{2015})
\end{bchapter}
\endbibitem

\bibitem[\protect\citeauthoryear{Li et~al.}{2014}]{li2014secrets}
\begin{bchapter}
\bauthor{\bsnm{Li}, \binits{Y.}},
\bauthor{\bsnm{Hou}, \binits{X.}},
\bauthor{\bsnm{Koch}, \binits{C.}},
\bauthor{\bsnm{Rehg}, \binits{J.M.}},
\bauthor{\bsnm{Yuille}, \binits{A.L.}}:
\bctitle{The secrets of salient object segmentation}.
In: \bbtitle{Proceedings of the IEEE Conference on Computer Vision and Pattern
  Recognition},
pp. \bfpage{280}--\blpage{287}
(\byear{2014})
\end{bchapter}
\endbibitem

\bibitem[\protect\citeauthoryear{Yang et~al.}{2013}]{yang2013saliency}
\begin{bchapter}
\bauthor{\bsnm{Yang}, \binits{C.}},
\bauthor{\bsnm{Zhang}, \binits{L.}},
\bauthor{\bsnm{Lu}, \binits{H.}},
\bauthor{\bsnm{Ruan}, \binits{X.}},
\bauthor{\bsnm{Yang}, \binits{M.-H.}}:
\bctitle{Saliency detection via graph-based manifold ranking}.
In: \bbtitle{Proceedings of the IEEE Conference on Computer Vision and Pattern
  Recognition},
pp. \bfpage{3166}--\blpage{3173}
(\byear{2013})
\end{bchapter}
\endbibitem

\bibitem[\protect\citeauthoryear{Fan et~al.}{2018}]{fan2018enhanced}
\begin{botherref}
\oauthor{\bsnm{Fan}, \binits{D.-P.}},
\oauthor{\bsnm{Gong}, \binits{C.}},
\oauthor{\bsnm{Cao}, \binits{Y.}},
\oauthor{\bsnm{Ren}, \binits{B.}},
\oauthor{\bsnm{Cheng}, \binits{M.-M.}},
\oauthor{\bsnm{Borji}, \binits{A.}}:
Enhanced-alignment measure for binary foreground map evaluation.
arXiv preprint arXiv:1805.10421
(2018)
\end{botherref}
\endbibitem

\bibitem[\protect\citeauthoryear{Fan et~al.}{2017}]{fan2017structure}
\begin{bchapter}
\bauthor{\bsnm{Fan}, \binits{D.-P.}},
\bauthor{\bsnm{Cheng}, \binits{M.-M.}},
\bauthor{\bsnm{Liu}, \binits{Y.}},
\bauthor{\bsnm{Li}, \binits{T.}},
\bauthor{\bsnm{Borji}, \binits{A.}}:
\bctitle{Structure-measure: A new way to evaluate foreground maps}.
In: \bbtitle{Proceedings of the IEEE International Conference on Computer
  Vision},
pp. \bfpage{4548}--\blpage{4557}
(\byear{2017})
\end{bchapter}
\endbibitem

\bibitem[\protect\citeauthoryear{Zhang et~al.}{2017a}]{zhang2017amulet}
\begin{bchapter}
\bauthor{\bsnm{Zhang}, \binits{P.}},
\bauthor{\bsnm{Wang}, \binits{D.}},
\bauthor{\bsnm{Lu}, \binits{H.}},
\bauthor{\bsnm{Wang}, \binits{H.}},
\bauthor{\bsnm{Ruan}, \binits{X.}}:
\bctitle{Amulet: Aggregating multi-level convolutional features for salient
  object detection}.
In: \bbtitle{Proceedings of the IEEE International Conference on Computer
  Vision},
pp. \bfpage{202}--\blpage{211}
(\byear{2017})
\end{bchapter}
\endbibitem

\bibitem[\protect\citeauthoryear{Zhang et~al.}{2017b}]{zhang2017learning}
\begin{bchapter}
\bauthor{\bsnm{Zhang}, \binits{P.}},
\bauthor{\bsnm{Wang}, \binits{D.}},
\bauthor{\bsnm{Lu}, \binits{H.}},
\bauthor{\bsnm{Wang}, \binits{H.}},
\bauthor{\bsnm{Yin}, \binits{B.}}:
\bctitle{Learning uncertain convolutional features for accurate saliency
  detection}.
In: \bbtitle{Proceedings of the IEEE International Conference on Computer
  Vision},
pp. \bfpage{212}--\blpage{221}
(\byear{2017})
\end{bchapter}
\endbibitem

\bibitem[\protect\citeauthoryear{Wu et~al.}{2019}]{wu2019cascaded}
\begin{bchapter}
\bauthor{\bsnm{Wu}, \binits{Z.}},
\bauthor{\bsnm{Su}, \binits{L.}},
\bauthor{\bsnm{Huang}, \binits{Q.}}:
\bctitle{Cascaded partial decoder for fast and accurate salient object
  detection}.
In: \bbtitle{Proceedings of the IEEE/CVF Conference on Computer Vision and
  Pattern Recognition},
pp. \bfpage{3907}--\blpage{3916}
(\byear{2019})
\end{bchapter}
\endbibitem

\bibitem[\protect\citeauthoryear{Qin et~al.}{2019}]{qin2019basnet}
\begin{bchapter}
\bauthor{\bsnm{Qin}, \binits{X.}},
\bauthor{\bsnm{Zhang}, \binits{Z.}},
\bauthor{\bsnm{Huang}, \binits{C.}},
\bauthor{\bsnm{Gao}, \binits{C.}},
\bauthor{\bsnm{Dehghan}, \binits{M.}},
\bauthor{\bsnm{Jagersand}, \binits{M.}}:
\bctitle{Basnet: Boundary-aware salient object detection}.
In: \bbtitle{Proceedings of the IEEE/CVF Conference on Computer Vision and
  Pattern Recognition},
pp. \bfpage{7479}--\blpage{7489}
(\byear{2019})
\end{bchapter}
\endbibitem

\bibitem[\protect\citeauthoryear{Zhao et~al.}{2020}]{zhao2020suppress}
\begin{bchapter}
\bauthor{\bsnm{Zhao}, \binits{X.}},
\bauthor{\bsnm{Pang}, \binits{Y.}},
\bauthor{\bsnm{Zhang}, \binits{L.}},
\bauthor{\bsnm{Lu}, \binits{H.}},
\bauthor{\bsnm{Zhang}, \binits{L.}}:
\bctitle{Suppress and balance: A simple gated network for salient object
  detection}.
In: \bbtitle{Computer vision--ECCV 2020: 16th European Conference, Glasgow, UK,
  August 23--28, 2020, Proceedings, Part II 16},
pp. \bfpage{35}--\blpage{51}
(\byear{2020}).
\bcomment{Springer}
\end{bchapter}
\endbibitem

\bibitem[\protect\citeauthoryear{Chen et~al.}{2018}]{chen2018reverse}
\begin{bchapter}
\bauthor{\bsnm{Chen}, \binits{S.}},
\bauthor{\bsnm{Tan}, \binits{X.}},
\bauthor{\bsnm{Wang}, \binits{B.}},
\bauthor{\bsnm{Hu}, \binits{X.}}:
\bctitle{Reverse attention for salient object detection}.
In: \bbtitle{Proceedings of the European Conference on Computer Vision (ECCV)},
pp. \bfpage{234}--\blpage{250}
(\byear{2018})
\end{bchapter}
\endbibitem

\bibitem[\protect\citeauthoryear{Song et~al.}{2023}]{song2023disentangle}
\begin{barticle}
\bauthor{\bsnm{Song}, \binits{Y.}},
\bauthor{\bsnm{Tang}, \binits{H.}},
\bauthor{\bsnm{Sebe}, \binits{N.}},
\bauthor{\bsnm{Wang}, \binits{W.}}:
\batitle{Disentangle saliency detection into cascaded detail modeling and body
  filling}.
\bjtitle{ACM Transactions on Multimedia Computing, Communications and
  Applications}
\bvolume{19}(\bissue{1}),
\bfpage{1}--\blpage{15}
(\byear{2023})
\end{barticle}
\endbibitem

\bibitem[\protect\citeauthoryear{Gao et~al.}{2020}]{gao2020highly}
\begin{bchapter}
\bauthor{\bsnm{Gao}, \binits{S.-H.}},
\bauthor{\bsnm{Tan}, \binits{Y.-Q.}},
\bauthor{\bsnm{Cheng}, \binits{M.-M.}},
\bauthor{\bsnm{Lu}, \binits{C.}},
\bauthor{\bsnm{Chen}, \binits{Y.}},
\bauthor{\bsnm{Yan}, \binits{S.}}:
\bctitle{Highly efficient salient object detection with 100k parameters}.
In: \bbtitle{European Conference on Computer Vision},
pp. \bfpage{702}--\blpage{721}
(\byear{2020}).
\bcomment{Springer}
\end{bchapter}
\endbibitem

\bibitem[\protect\citeauthoryear{Shen et~al.}{2022}]{shen2022fully}
\begin{barticle}
\bauthor{\bsnm{Shen}, \binits{K.}},
\bauthor{\bsnm{Zhou}, \binits{X.}},
\bauthor{\bsnm{Wan}, \binits{B.}},
\bauthor{\bsnm{Shi}, \binits{R.}},
\bauthor{\bsnm{Zhang}, \binits{J.}}:
\batitle{Fully squeezed multiscale inference network for fast and accurate
  saliency detection in optical remote-sensing images}.
\bjtitle{IEEE Geoscience and Remote Sensing Letters}
\bvolume{19},
\bfpage{1}--\blpage{5}
(\byear{2022})
\end{barticle}
\endbibitem

\bibitem[\protect\citeauthoryear{Li et~al.}{2021}]{li2021multi}
\begin{barticle}
\bauthor{\bsnm{Li}, \binits{G.}},
\bauthor{\bsnm{Liu}, \binits{Z.}},
\bauthor{\bsnm{Lin}, \binits{W.}},
\bauthor{\bsnm{Ling}, \binits{H.}}:
\batitle{Multi-content complementation network for salient object detection in
  optical remote sensing images}.
\bjtitle{IEEE Transactions on Geoscience and Remote Sensing}
\bvolume{60},
\bfpage{1}--\blpage{13}
(\byear{2021})
\end{barticle}
\endbibitem

\bibitem[\protect\citeauthoryear{Krizhevsky
  et~al.}{2009}]{krizhevsky2009learning}
\begin{botherref}
\oauthor{\bsnm{Krizhevsky}, \binits{A.}},
\oauthor{\bsnm{Hinton}, \binits{G.}}, et al.:
Learning multiple layers of features from tiny images
(2009)
\end{botherref}
\endbibitem

\bibitem[\protect\citeauthoryear{Sandler et~al.}{2018}]{sandler2018mobilenetv2}
\begin{bchapter}
\bauthor{\bsnm{Sandler}, \binits{M.}},
\bauthor{\bsnm{Howard}, \binits{A.}},
\bauthor{\bsnm{Zhu}, \binits{M.}},
\bauthor{\bsnm{Zhmoginov}, \binits{A.}},
\bauthor{\bsnm{Chen}, \binits{L.-C.}}:
\bctitle{Mobilenetv2: Inverted residuals and linear bottlenecks}.
In: \bbtitle{Proceedings of the IEEE Conference on Computer Vision and Pattern
  Recognition},
pp. \bfpage{4510}--\blpage{4520}
(\byear{2018})
\end{bchapter}
\endbibitem

\bibitem[\protect\citeauthoryear{Zhang et~al.}{2018}]{zhang2018shufflenet}
\begin{bchapter}
\bauthor{\bsnm{Zhang}, \binits{X.}},
\bauthor{\bsnm{Zhou}, \binits{X.}},
\bauthor{\bsnm{Lin}, \binits{M.}},
\bauthor{\bsnm{Sun}, \binits{J.}}:
\bctitle{Shufflenet: An extremely efficient convolutional neural network for
  mobile devices}.
In: \bbtitle{Proceedings of the IEEE Conference on Computer Vision and Pattern
  Recognition},
pp. \bfpage{6848}--\blpage{6856}
(\byear{2018})
\end{bchapter}
\endbibitem

\bibitem[\protect\citeauthoryear{Ma et~al.}{2018}]{ma2018shufflenet}
\begin{bchapter}
\bauthor{\bsnm{Ma}, \binits{N.}},
\bauthor{\bsnm{Zhang}, \binits{X.}},
\bauthor{\bsnm{Zheng}, \binits{H.-T.}},
\bauthor{\bsnm{Sun}, \binits{J.}}:
\bctitle{Shufflenet v2: Practical guidelines for efficient cnn architecture
  design}.
In: \bbtitle{Proceedings of the European Conference on Computer Vision (ECCV)},
pp. \bfpage{116}--\blpage{131}
(\byear{2018})
\end{bchapter}
\endbibitem

\bibitem[\protect\citeauthoryear{Iandola et~al.}{2016}]{iandola2016squeezenet}
\begin{botherref}
\oauthor{\bsnm{Iandola}, \binits{F.N.}},
\oauthor{\bsnm{Han}, \binits{S.}},
\oauthor{\bsnm{Moskewicz}, \binits{M.W.}},
\oauthor{\bsnm{Ashraf}, \binits{K.}},
\oauthor{\bsnm{Dally}, \binits{W.J.}},
\oauthor{\bsnm{Keutzer}, \binits{K.}}:
Squeezenet: Alexnet-level accuracy with 50x fewer parameters and< 0.5 mb model
  size.
arXiv preprint arXiv:1602.07360
(2016)
\end{botherref}
\endbibitem

\end{thebibliography}

\end{document}